\documentclass[letterpaper]{article}
\usepackage{aaai24}
\usepackage{times}
\usepackage{helvet}
\usepackage{courier}
\usepackage[hyphens]{url}
\usepackage{graphicx}
\usepackage{natbib}
\usepackage{caption}
\usepackage{algorithm}
\usepackage{algorithmic}
\usepackage{newfloat}
\usepackage{listings}

\usepackage{url}
\usepackage{booktabs}
\usepackage{amsfonts}
\usepackage{nicefrac}
\usepackage{microtype}
\usepackage{xcolor}
\usepackage{subfiles}

\usepackage{graphicx}
\graphicspath{ {./images/} }
\usepackage{comment}
\usepackage{csquotes}

\usepackage{tikz}
\usepackage{pgfplots}
\usepackage{pgfplotstable}
\usetikzlibrary{patterns}
\usepgfplotslibrary{colormaps}
\pgfplotsset{compat=1.18}
\usetikzlibrary{pgfplots.colormaps}
\usepackage{pgf-pie} 
\usepackage{xcolor,colortbl}
\usepackage{multirow}
\usepackage{xurl}
\usepackage{siunitx}

\usepackage{dcolumn}
\usepackage{makecell}

\usetikzlibrary{pgfplots.statistics}
\usetikzlibrary{matrix}
\usetikzlibrary{calc}

\newcolumntype{d}[1]{D{.}{.}{#1}}
\newcommand{\mc}[1]{\multicolumn{1}{c@{}}{#1}}
\newcommand{\gray}[1]{%
{\cellcolor{gray!\number\numexpr7*#1/12\relax}}%
}

\pgfplotstableread{
OpenAI	Scaling	FaceCLIP	Human
-0.12	0.13	-0.04	0.76
0.09	0.42	0.27	0.73
0.08	0.05	-0.02	0.79
0.08	0.37	0.13	0.65
0.75	0.80	0.81	0.98
0.73	0.85	0.83	0.97
0.61	0.67	0.67	0.83
0.21	0.11	0.20	0.53
0.81	0.84	0.90	0.94
0.15	0.07	0.26	0.28
0.30	0.23	0.06	0.82
0.03	0.03	-0.06	0.44
0.14	0.07	0.14	0.78
0.36	0.55	0.48	0.96
0.09	0.22	-0.01	0.68
0.30	0.12	0.26	0.72
0.02	0.33	0.14	0.92
0.51	0.60	0.59	0.96
-0.21	-0.21	-0.27	0.58
0.42	0.58	0.51	0.87
0.24	0.19	0.26	0.68
0.25	0.20	0.33	0.67
0.58	0.78	0.72	0.95
0.47	0.66	0.64	0.88
0.55	0.77	0.69	0.87
0.53	0.71	0.65	0.89
0.37	0.63	0.40	0.87
0.26	0.38	0.40	0.93
0.28	0.03	0.30	0.95
-0.21	-0.11	-0.23	0.27
0.19	0.40	0.29	0.73
0.60	0.42	0.60	0.87
0.01	0.23	-0.03	0.49
0.60	0.75	0.75	0.94
}\datatable

\pgfplotstableread{
80m	400m	2b
0.6335174953959485	0.6568265682656826	0.6648351648351648
0.6487985212569316	0.6204379562043796	0.6763636363636364
0.6771217712177122	0.6611721611721612	0.704119850187266
0.6356877323420075	0.6850828729281768	0.6654135338345865
0.6528028933092225	0.6900369003690037	0.6611418047882136
0.6363636363636364	0.6943396226415094	0.6611418047882136
0.6570397111913358	0.697196261682243	0.6788990825688074
0.5981651376146789	0.663003663003663	0.6559546313799622
0.649090909090909	0.6330275229357798	0.7066420664206642
}\jaccarddata

\pgfplotstableread{
    Group,Value1,Error1,Value2,Error2,Value3,Error3,
    Group1,35,5,40,6,45,7,
    Group2,45,7,50,8,55,9,
    Group3,60,8,65,9,70,10,
}\famtable

\DeclareCaptionStyle{ruled}{labelfont=normalfont,labelsep=colon,strut=off} 
\floatstyle{ruled}
\newfloat{listing}{tb}{lst}{}
\floatname{listing}{Listing}

\title{Dataset Scale and Societal Consistency Mediate Facial Impression Bias in Vision-Language AI}

\author {
    Robert Wolfe,
    Aayushi Dangol,
    Alexis Hiniker,
    Bill Howe
}
\affiliations {
    University of Washington\\
    rwolfe3@uw.edu, adango@uw.edu, alexisr@uw.edu, billhowe@uw.edu
}

\begin{document}

\maketitle

\begin{abstract}
Multimodal AI models capable of associating images and text hold promise for numerous domains, ranging from automated image captioning to accessibility applications for blind and low-vision users. However, uncertainty about bias has in some cases limited their adoption and availability. In the present work, we study 43 CLIP vision-language models to determine whether they learn human-like facial impression biases, and we find evidence that such biases are reflected across three distinct CLIP model families. We show for the first time that the the \textit{degree} to which a bias is shared across a society predicts the degree to which it is reflected in a CLIP model. Human-like impressions of visually \textit{unobservable} attributes, like trustworthiness and sexuality, emerge only in models trained on the largest dataset, indicating that a better fit to uncurated cultural data results in the reproduction of increasingly subtle social biases. Moreover, we use a hierarchical clustering approach to show that dataset size predicts the extent to which the underlying structure of facial impression bias resembles that of facial impression bias in humans. Finally, we show that Stable Diffusion models employing CLIP as a text encoder learn facial impression biases, and that these biases intersect with racial biases in Stable Diffusion XL-Turbo. While pretrained CLIP models may prove useful for scientific studies of bias, they will also require significant dataset curation when intended for use as general-purpose models in a zero-shot setting.
\end{abstract}
\section{Introduction}

OpenAI's multimodal GPT-4 powers the beta version of Be My AI, an extension of the Be My Eyes app \cite{bemyai,bemyeyes} that provides "instantaneous identification, interpretation, and conversational visual assistance" to blind and low-vision users. Until recently, the app allowed users to ask questions about images of people and receive live explanations. The temporary discontinuation of this feature was motivated by concern that GPT-4 ``would say things it shouldn’t about people’s faces, such as assessing their gender or emotional state'' \cite{hill2023openai}.

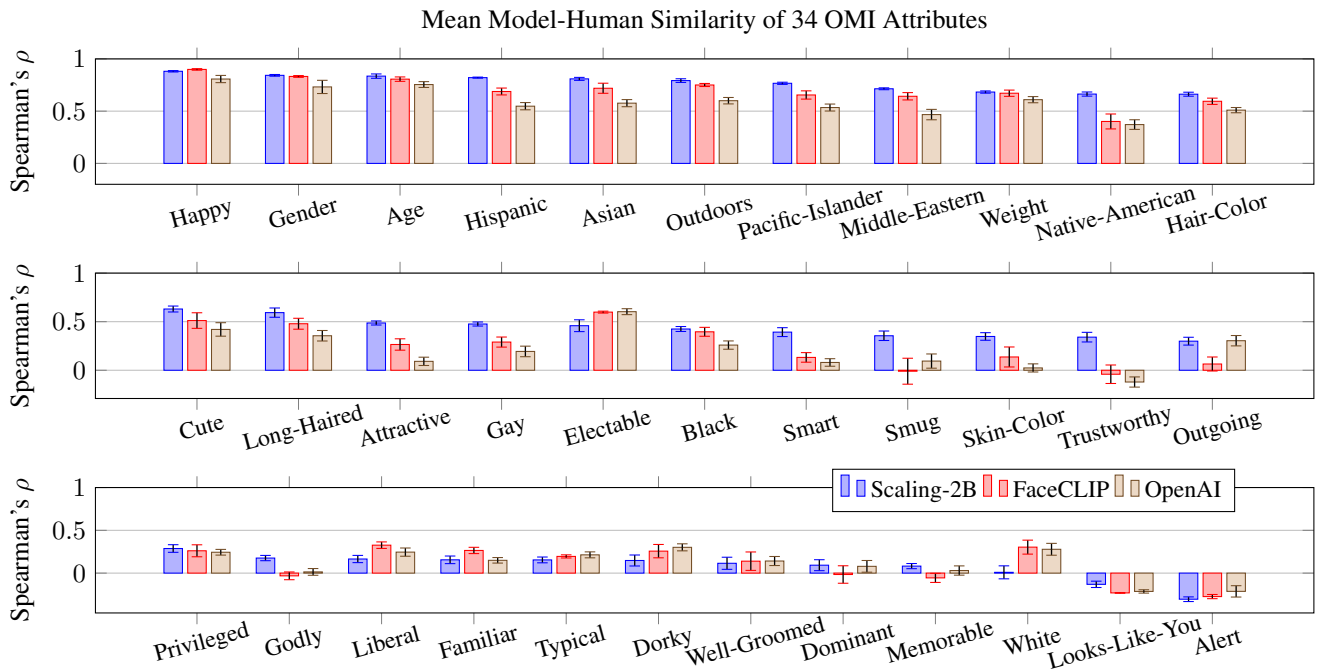
\begin{figure*}
    \centering
    \begin{tikzpicture}
        \begin{axis}[
            width=\textwidth,
            height=3.25cm,
            ybar,
            bar width=7pt,
            xtick=data,
            ymax=1,
            ymin=-.2,
            ymajorgrids=true,
            xticklabels from table={familydata.txt}{Attribute},
            xticklabel style={font=\small,rotate=15, anchor=north},
            ylabel={Spearman's $\rho$},
            title={Mean Model-Human Similarity of 34 OMI Attributes}
            ]
            \addplot+[error bars/.cd,
                      y dir=both,
                      y explicit,
                      error mark=-,
                      error bar style={
            draw=black,
            line width=.5pt,}]
            table[x expr=\coordindex, y=Scaling_2B, y error=Scaling_2B_Error] {familydata.txt};

            \addplot+[error bars/.cd,
                      y dir=both,
                      y explicit,
                      error mark=-,
                      error bar style={
            draw=black,
            line width=.5pt,}]
            table[x expr=\coordindex, y=FaceCLIP, y error=FaceCLIP_Error] {familydata.txt};
            
            \addplot+[error bars/.cd,
                      y dir=both,
                      y explicit,
                      error mark=-,
                      error bar style={
            draw=black,
            line width=.5pt,}]
            table[x expr=\coordindex, y=OpenAI, y error=OpenAI_Error] {familydata.txt};
        
        \end{axis}
    \end{tikzpicture}
    \begin{tikzpicture}
        \begin{axis}[
            width=\textwidth,
            height=3.25cm,
            ymajorgrids=true,
            ybar,
            bar width=7pt,
            xtick=data,
            ymax=1,
            xticklabels from table={familydata2.txt}{Attribute},
            xticklabel style={font=\small,rotate=15, anchor=north},
            legend style={at={(0.5,-0.15)},
                anchor=north,legend columns=-1},
            ylabel={Spearman's $\rho$},
            ]

            \addplot+[error bars/.cd,
                      y dir=both,
                      y explicit,
                      error mark=-,
                      error bar style={
            draw=black,
            line width=.5pt,}]
            table[x expr=\coordindex, y=Scaling_2B, y error=Scaling_2B_Error] {familydata2.txt};

            \addplot+[error bars/.cd,
                      y dir=both,
                      y explicit,
                      error mark=-,
                      error bar style={
            draw=black,
            line width=.5pt,}]
            table[x expr=\coordindex, y=FaceCLIP, y error=FaceCLIP_Error] {familydata2.txt};

            \addplot+[error bars/.cd,
                      y dir=both,
                      y explicit,
                      error mark=-,
                      error bar style={
            draw=black,
            line width=.5pt,}]
            table[x expr=\coordindex, y=OpenAI, y error=OpenAI_Error] {familydata2.txt};
        \end{axis}
    \end{tikzpicture}
    \begin{tikzpicture}
        \begin{axis}[
            width=\textwidth,
            height=3.25cm,
            ybar,
            ymajorgrids=true,
            bar width=7pt,
            xtick=data,
            ymax=1,
            xticklabels from table={familydata3.txt}{Attribute},
            xticklabel style={font=\small,rotate=15, anchor=north},
            legend style={at={(0.5,-0.15)},
                anchor=north,legend columns=-1},
            ylabel={Spearman's $\rho$},
            legend style={at={(0.77,1.15)},
                anchor=north,legend columns=-1,font=\small}
            ]
            \addplot+[error bars/.cd,
                      y dir=both,
                      y explicit,
                      error mark=-,
                      error bar style={
            draw=black,
            line width=.5pt,}]
            table[x expr=\coordindex, y=Scaling_2B, y error=Scaling_2B_Error] {familydata3.txt};

            \addplot+[error bars/.cd,
                      y dir=both,
                      y explicit,
                      error mark=-,
                      error bar style={
            draw=black,
            line width=.5pt,}]
            table[x expr=\coordindex, y=FaceCLIP, y error=FaceCLIP_Error] {familydata3.txt};
            
            \addplot+[error bars/.cd,
                      y dir=both,
                      y explicit,
                      error mark=-,
                      error bar style={
            draw=black,
            line width=.5pt,}]
            table[x expr=\coordindex, y=OpenAI, y error=OpenAI_Error] {familydata3.txt};
        \addlegendentry{Scaling-2B}
        \addlegendentry{FaceCLIP}
        \addlegendentry{OpenAI}
        
        \end{axis}
    \end{tikzpicture}

    \caption{\small CLIP models learn human-like facial impression biases. The highest model-human correlations are obtained for intuitively visual categories that are broadly shared by a society (such as gender, age, and happiness). Models trained on the largest dataset (LAION-2B) exhibit more human-like biases than FaceCLIP or OpenAI models for most attributes.}
    \label{fig:model_human_similarity_bar_chart}
\end{figure*}

The decision belies a broader concern: that by learning to associate language and images, multimodal AI may make insufficiently informed judgments about human attributes based solely on a person's face. When studied in human subjects, this kind of inference is known as a ``first impression'' or ``facial impression'' bias \cite{todorov2017face}, and it is known to affect consequential spheres of human social life such as criminal sentencing \cite{wilson2015facial}, employment decisions \cite{stoker2016facial}, and political elections \cite{antonakis2009predicting}. Such impressions can include traits like trustworthiness, which are unobservable from a person's face and societally mediated to extent that they are consistent in a population \cite{todorov2017face}. While psychologists have used computational geometry and supervised machine learning approaches to modeling facial impression biases \cite{blanz2023morphable}, it is not known whether semi-supervised vision-language AI models could inadvertently learn such biases in pretraining and propagate them to the many domains in which such models are used.

While features permitting facial image analysis are disabled in GPT-4, the opportunity to study facial impression bias is afforded by CLIP (``Contrastive Language Image Pretraining''), a state-of-the-art vision-language model that allows users to define text classes at inference using natural language \cite{radford2021learning}. Rather than fine-tuning CLIP to model facial impressions similar to prior work using supervised learning, we study this bias in three families of pretrained CLIP models used in a wide range of multimodal computer vision tasks: the nine models trained by OpenAI \cite{radford2021learning}; five ``FaceCLIP'' models post-trained for facial analysis \cite{zheng2022general}; and 29 ``Scaling'' models trained by \citet{cherti2022reproducible} on systematically differing amounts of data, allowing for statistical analysis of the effects of model and dataset parameters on facial impression bias.

Analyzing whether CLIP models learn human-like facial impression biases requires a reliable source of human data. This research uses the authoritative One Million Impressions (OMI) dataset of \citet{peterson2022deep}, which includes 1,004 images of faces rated by human participants across 34 attributes, with which \citet{peterson2022deep} learned a supervised model of facial impression biases. In the present work, we used CLIP to compute the similarity of each OMI image to text prompts for the 34 attributes, mimicking the task given to human subjects, and we compared the CLIP similarities to human subject ratings. We offer four primary findings:

\begin{enumerate}

\item \textbf{CLIP models learn societal facial impression biases, including for unobservable traits such as trustworthiness and sexuality.} Moreover, the extent to which an attribute bias is learned by a CLIP model is strongly correlated with the inter-rater reliability (IRR) of human judgments of the attribute (Spearman's $\rho=.73$ for OpenAI models; $\rho=.76$ for FaceCLIP models; and $\rho=.72$ for Scaling models). A multiple linear regression predicting the similarity of CLIP bias to Human bias finds that the IRR of the attribute plays a larger role than any model-related variable, with $t(912) = 25.47$, $p<.001$. The extent to which a facial impression bias is learned by a model depends on how consistently it is shared in the population that produced the data on which the models trains.

\item \textbf{Dataset Scale is a significant predictor of facial impression bias in CLIP.} Comparison of model-human similarity in two groups of nine CLIP models trained on LAION-80M (80 million examples) and LAION-400M (407 million examples) yields large effect sizes ($d>0.8$) and statistically significant ($p<.05$) paired samples $t$-tests for 17 of 34 attributes, indicating increases in the human similarity of bias in models trained on LAION-400M. Differences between models trained on LAION-2B (2.32 billion examples) and LAION-400M are mostly not significant, with the notable exception of \textit{unobservable} attributes like trustworthiness ($d=1.33$, $p<.05$) and sexuality ($d=1.14$, $p<.05$). While models trained on larger datasets exhibit stronger task performance \cite{cherti2022reproducible}, they also more faithfully reflect the biases of the population that produced the data.

\item \textbf{CLIP models learn human-like associations \textit{between} facial impression biases.} Hierarchical clustering of CLIP and OMI attribute correlation matrices reveals similar groupings of traits, including clusters related to ethnicity, and clusters grouping gender, sexuality, and age. Computing the normalized Frobenius inner product of CLIP correlation matrices with the OMI matrix reveals increasing similarity as pretraining data size increases, with a one-way ANOVA yielding $F(2) = 15.71$, $p<.001$, and large ($d>0.8$) pairwise effect sizes between groups of Scaling-80M with 400M and 2B models.

\item \textbf{Stable Diffusion text-to-image models employing CLIP as a text encoder learn facial impression biases that intersect with demographic biases.} Images generated by Stable Diffusion XL-Turbo (SDXL) are classified by a classifier fit using the OMI images. Classifications reflect human facial impression biases for subjective, observable attributes like attractiveness (F1=.98), and to a lesser extent for unobservable attributes like liberal (F1=.68) and smart (F1=.65). Applying the classifier to SDXL images generated for White and Black prompts reveals biases in SDXL differentially associating White individuals with traits like memorable, attractive, electable, and happy.

\end{enumerate}

\noindent Training vision-language models on vast web-scraped datasets produces subtle yet consequential emergent biases. Such models may serve as useful tools for social science, potentially illuminating factors that contribute to human bias. However, the presence of such biases in vision-language AI also renders fraught these models' real-world use, as they may subtly reinforce existing inequities in an online environment increasingly mediated by AI. Code and data are available at https://github.com/wolferobert3/vl-facial-impressions-bias.
\section{Related Work}

We review the related work on facial impression bias, vision-language AI, and the impact of scale in deep learning.

\subsection{Facial Impression Bias}

A wealth of psychological research indicates that humans make immediate judgments about the attributes of people they do not know based solely on facial appearance \cite{willis2006first, oh2019revealing,charlesworth2019children}. Information inferred from faces includes character traits (like trustworthiness and outgoingness) and socially constructed group memberships (like gender and ethnicity) as well as relatively objective traits (like hair color and weight) \cite{todorov2017face,peterson2022deep}. Research on first-impression biases in humans has found that the inference of attributes from facial appearance plays a role in numerous consequential domains, including employment decisions \cite{stoker2016facial,graham2017corporate,swider2021first}, criminal sentencing \cite{wilson2015facial,johnson2017facial}, and the election of political candidates \cite{antonakis2009predicting,olivola2010elected,lenz2011looking,jackle2020catwalk}. While facial impression biases may be consistent among a population, inferences of unobservable attributes such as character traits are inaccurate and often reflect societal stereotypes \cite{sutherland2022understanding, todorov2017face}. AI systems are increasingly employed to automate or mediate access to information in domains such as hiring \cite{li2021algorithmic}, political analysis and advertising \cite{papakyriakopoulos2022algorithms}, and law \cite{choi2023chatgpt}, and to the extent that such systems reflect facial impression biases, they may have socially undesirable impacts.

\subsection{Relationship to Social Group Biases} Some studies suggest a connection between first-impression biases and demographic traits such as gender and ethnicity. \citet{oh2019revealing} find that gender biases associating men with competence are reflected in participant impressions of the competence of faces. \citet{xie2021facial} find that the structure of impressions of novel faces is predicted by learned social stereotypes about gender and race. \citet{peterson2022deep} find that facial impression biases are correlated with demographics, such that judgments of traits like ``cuteness'' are related to age. The relationship between first-impression bias and social stereotypes can have real-world consequences. For example, White phenotypic prototypicality (looking like the average White person) can moderate use of force by police \cite{kahn2016protecting}.

\subsection{Computational Models of Facial Impression Bias}

\citet{peterson2022deep} use the OMI dataset to model facial impressions using the StyleGAN-2 network \cite{karras2019style}, and demonstrate its capacity to manipulate faces such that the average U.S. perceiver would consider them similar to an attribute (such as trustworthiness). They build on research on the scientific modeling of facial impression biases, which commonly utilizes techniques including landmark annotations of faces \cite{turk1991eigenfaces}, parametric three-dimensional mesh modeling \cite{blanz2023morphable}, geometric morphological analysis \cite{sano2023computational}, and supervised deep learning models  \cite{yu2022deep}. As noted by \citet{peterson2022deep}, creating a computational model of a \textit{bias} differs from modeling the attribute itself (\textit{i.e.,} trying to predict if an individual is trustworthy from their face, rather than whether the average person would \textit{perceive} an individual as trustworthy), which would amount to physiognomy \cite{y2023physiognomy} for an unobservable attribute like trustworthiness.

\subsection{CLIP and Vision-Language AI}

The present work studies CLIP, a multimodal vision-language model pretrained using a symmetric cross-entropy loss \cite{oord2018representation,zhang2020contrastive} to pair images with associated text captions \cite{radford2021learning}. After pretraining, CLIP can rank, retrieve, or classify images based on association with text classes specified at inference rather than pre-selected at the time of training, making it a ``zero-shot'' vision-language model \cite{radford2021learning}, as well as a good source for semantically rich embeddings \cite{wolfe2022contrastive}. CLIP is composed of a language model (usually GPT2 \cite{radford2019language}), and an image encoder, such as a Vision Transformer (``ViT'') \cite{dosovitskiy2020image} or a ResNet \cite{he2016deep}. The language and image models are jointly pretrained, and representations are projected into a multimodal embedding space, in which cosine similarity quantifies the similarity between image and text \cite{radford2021learning}. In addition to standard CLIP models, we study ``FaceCLIP'' models trained by \citet{zheng2022general}, who introduce Facial Representation Learning (FaRL), which combines CLIP training with a masked image modeling objective \cite{xie2022simmim} and trains on a faces-only subset of LAION-400M. Models trained using FaRL set state-of-the-art on downstream facial analysis tasks such as face parsing \cite{zheng2022general}. \looseness=-1

\subsection{Text-to-Image Generators}

CLIP is an essential component for many generative text-to-image models. One of the first uses of a CLIP model was to provide training supervision to OpenAI's first DALL-E image generator model \cite{ramesh2021zero}. Other text-to-image generators like VQGAN-CLIP similarly use CLIP embedding space measurements in their objective function \cite{crowson2022vqgan}. More recent image generators such as Stable Diffusion 2 employ CLIP models as text encoders \cite{Rombach_2022_CVPR}, passing CLIP text embeddings to a U-Net or similar latent diffusion architecture capable of generating an image conditioned on those text embeddings. More recently, DALL-E 3 (``unCLIP'') decodes images directly from a CLIP embedding space, translating CLIP text embeddings into image embeddings, and inverting them \cite{ramesh2022hierarchical}. As discussed in the Data section, we study open-weight text-to-image generators utilizing a CLIP text encoder.

\subsection{Impact of Scale in Deep Learning and in CLIP}

Research shows that the impact of data scale on deep learning models is empirically predictable \cite{hestness2017deep} and that task performance scales with training dataset size \cite{sun2017revisiting, brown2020language}. \citet{zhai2022scaling} empirically demonstrate that both model and data scale impact visual task performance, and set new state of the art on Imagenet \cite{deng2009imagenet} by efficiently scaling a ViT. In CLIP models, \citet{cherti2022reproducible} demonstrate a relationship between pretraining data scale and task performance. Prior work also demonstrates increases in hate speech in CLIP models trained on larger uncurated datasets \cite{birhane2024dark}.

\subsection{Bias in Vision-Language AI}

Prior research identifies societal biases in CLIP models, which have been found to over-represent images of men in retrieval contexts \cite{wang2021gender}, to exhibit social stereotypes when used in robotics applications \cite{hundt2022robots}, and to reflect biased cultural defaults related to race and nationality \cite{wolfe2022american,wolfe2022markedness}. Generative text to image models typically reflect biases of the CLIP model from which they learn or decode \cite{luccioni2023stable,bianchi2023easily}, as do vision-language chatbots that use CLIP as an image encoder \cite{fraser2024examining}. As with other machine learning models \cite{hall2022systematic}, bias in CLIP can be traced back to its pretraining data, and audits of LAION-400M have found examples of racial and gender biases \cite{birhane2021multimodal}.

\subsection{Synthetic Media for Vision-Language Research}

Our work builds on prior work employing synthetic images to study societal biases. Similar to the OMI dataset studied in this research, \citet{wolfe2022hypodescent} assessed classification biases in CLIP related to images of multiracial individuals by generating images of thousands of faces using StyleGAN-2. More recently, \citet{fraser2024examining} use Midjourney \cite{midjourney} to generate pairs of images of Black and White or Male and Female individuals in identical attire and situations, which they use to assess bias in conversational vision-language models such as Instruct-BLIP \cite{instructblip}. The use of synthetic data is well-motivated for bias research, where it may be undesirable to assign values related to traits like trustworthiness or smugness to the faces of human subjects in a dataset intended for public release.
\section{Data}

This research uses the One Million Impressions (OMI) dataset and 43 English-language CLIP models trained on web-scraped text-and-image datasets, as well as three text-to-image generators employing CLIP models as text encoders.

\subsection{The One Million Impressions Dataset}

The OMI dataset is a collection of 1,004 images of human faces produced by \citet{peterson2022deep} using StyleGAN-2 \cite{Karras2019stylegan2}. Each face is rated by 30 or more human participants on Amazon Mechanical Turk for 34 attributes. For each attribute, participants rate the face on a sliding scale, where one end represents one pole of an attribute binary (such as ``trustworthy'') and the other end represents the opposing pole of the binary (such as ``untrustworthy''). The OMI dataset records the mean participant rating for each of the 34 attributes. Consistent with \citet{peterson2022deep}, we use these ratings as measurements of human bias at a societal scale, against which CLIP associations can be compared.

\begin{figure}
    \centering
    \includegraphics[width=.7\linewidth, trim={0 9cm 0 0},clip]{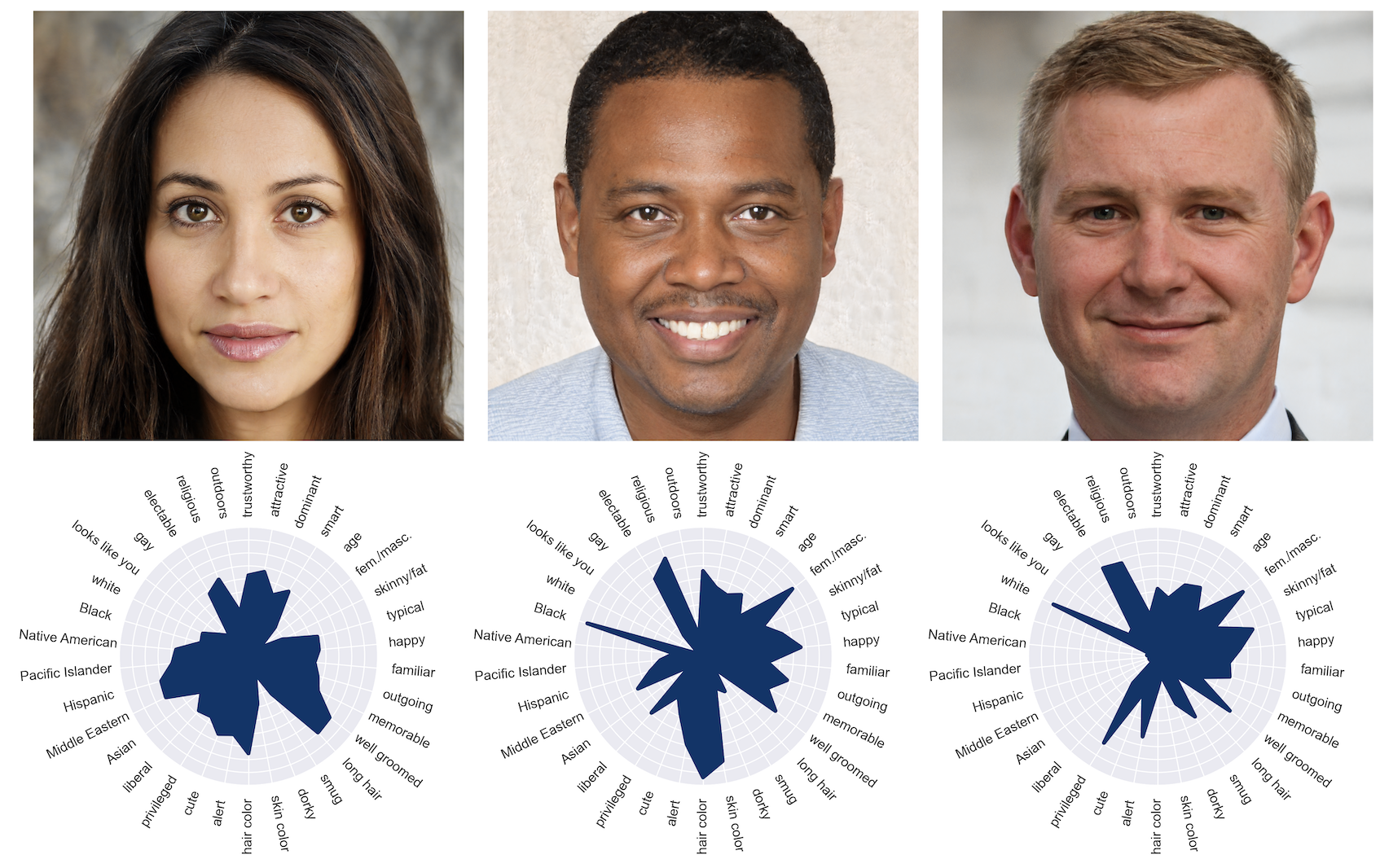}
    \caption{\small Examples from the OMI dataset repository at https://github.com/jcpeterson/omi, used as stimuli in our research.}
    \label{fig:omi_examples}
\end{figure}

\subsection{CLIP Training Data}

We study CLIP models pretrained on one of five datasets, ordered from smallest to largest:

\begin{itemize}
\item \textbf{LAION-Face}: A 20-million sample subset of human faces and captions filtered from LAION-400M (see below) using RetinaFace \cite{deng2019retinaface} and intended for training facial analysis models \cite{zheng2022general}.

\item \textbf{LAION-80M}: An 80-million sample subset of LAION-2B (see below) created by \citet{cherti2022reproducible} to study scaling behavior in CLIP.

\item \textbf{LAION-Aesthetics}: A 120-million sample subset of aesthetically pleasing images from LAION-5B as determined using a CLIP model \cite{schuhmann2022laion}.

\item \textbf{WebImageText (WIT)}: A web-scraped corpus of 400 million images and captions, constructed by \citet{radford2021learning} from a query list using  Wikipedia and WordNet.

\item \textbf{LAION-400M} An open source collection of 407 million image-text pairs intended to replicate the WIT dataset \cite{schuhmann2021laion}.

\item \textbf{LAION-2B}: An open source English-language dataset of 2.32 billion image-text pairs \cite{schuhmann2022laion}. 
\end{itemize}

\subsection{Pretrained CLIP Models}

This research studies the following CLIP models:

\begin{itemize}
\item \textbf{OpenAI CLIP}: 9 models pretrained by \citet{radford2021learning} on the WIT dataset.

\item \textbf{Scaling CLIP}: 29 models pretrained by \citet{cherti2022reproducible} on LAION-80M, LAION-400M, and LAION-2B to study CLIP scaling behavior. 

\item \textbf{FaceCLIP}: 5 models trained on the LAION-Face dataset of \citet{zheng2022general}. FaceCLIP models post-train from pretrained OpenAI CLIP-ViT models.
\end{itemize}

\subsection{Pretrained Stable Diffusion Models}

This research studies three Stable Diffusion (SD) models:

\begin{itemize}
    \item \textbf{Stable Diffusion XL-Turbo}: A high-resolution text-to-image generator employing adversarial distillation diffusion to speed up the rate of image generation \cite{sauer2023adversarial}. Uses a CLIP-ViT-L and CLIP-ViT-bigG for its text encoder, and pretrains on an internal dataset.
    \item \textbf{Runway Stable Diffusion 1.5}: A high-resolution text-to-image generator finetuned on LAION-Aesthetics \cite{Rombach_2022_CVPR}. Uses a CLIP-ViT-L-14 as the text encoder, and pretrains on LAION-5B.
    \item \textbf{Stable Diffusion 2}: A text-to-image generator using a CLIP-ViT-H as the text encoder, and pretraining on a filtered subset of LAION-5B \cite{Rombach_2022_CVPR}.
\end{itemize}
\section{Approach}

We used embeddings from 43 CLIP models to compare facial impression biases measured in CLIP to biases measured in humans by Peterson et al. \cite{peterson2022deep}, and extended subspace projection methods from prior work to study bias in generative text-to-image models.

\subsection{Obtaining Image and Text Embeddings}

We obtained image embeddings for the 1,004 images in the OMI dataset after projection to each CLIP model's text-image latent space. Text embeddings use the ``a photo of \textit{image class}'' prompt recommended by \citet{radford2021learning}. Because OMI consists of images of faces, we modify this prompt to ``a photo of someone who is \textit{attribute}.'' In keeping with the binary sliding scale of \citet{peterson2022deep}, we computed an image's association with each attribute by subtracting its cosine similarity with one pole of the attribute binary (``a photo of someone who has dark hair'') from its similarity with opposing pole (``a photo of someone who has light hair''). Formally, given a model $m$ from which embeddings are obtained, the association $m^{a}_{j}$ of an image vector $\vec{i_{j}}$ at index $j$ of the OMI dataset with an attribute $a$ is the difference of the vector's cosine similarity with a positive pole text vector $\vec{t_{a^{+}}}$ and its cosine similarity with a negative pole text vector $\vec{t_{a^{-}}}$:

\begin{equation}
    m^{a}_{j} = cos(\vec{i_{j}}, \vec{t}_{a^{+}}) - cos(\vec{i_{j}}, \vec{t}_{a^{-}})
\end{equation}

\subsection{Adjusting Prompts for Negation}

CLIP may fail to adjust for negation in text prompts \cite{parcalabescu2021valse} and can behave like a visual bag-of-words model \cite{yuksekgonul2022and}. For example, CLIP might match the text ``a photo with no apples'' to a photo of apples, due to how unlikely it is for a text caption (\textit{i.e.,} CLIP's training supervision) to describe something not present in the photo. To adjust for this, negative pole prompts were chosen such that they did not simply negate the positive  class. For example, the ``outgoing'' attribute uses ``a photo of someone who is shy'' as the negative text class, rather than ``a photo of someone who is not outgoing.'' This strategy is not viable for some attributes, like those related to ethnicity, which instead use ``a photo of someone'' as the negative prompt. The full set of text prompts is provided in the appendix. \looseness=-1

\subsection{Computing CLIP Model-Human Similarity}

We denote the ordered set of $n$=1,004 OMI images as $I$. The vector of associations $\mathbf{m^{a}}$ for a model $m$ with an attribute $a$ for all images $i \in I$ is given by:

\begin{equation}
\mathbf{m^{a}} = (m^{a}_{0}, m^{a}_{1}, \ldots, m^{a}_{n-1}, m^{a}_{n})
\end{equation}

\noindent Similarly, the vector of human-rated associations $\mathbf{h^{a}}$ for attribute $a$ for all images $i \in I$ is given by:

\begin{equation}
\mathbf{h^{a}} = (h^{a}_{0}, h^{a}_{1}, \ldots, h^{a}_{n-1}, h^{a}_{n})
\end{equation}

\noindent where $h^{a}_{j}$ denotes the OMI mean for image $\vec{i}_{j}$ at index $j$. The similarity $s^{a}_{m}$ of bias in a model $m$ for attribute $a$ to human bias is given by Spearman's $\rho$:

\begin{equation}
    s^{a}_{m} = \rho(\mathbf{m^{a}},\mathbf{h^{a}})
\end{equation}

\subsection{CAT: Correlated Attribute Test}

We compute the correlation between two attributes in a CLIP model using a simple test we call the CAT. As above, the vector of associations $\mathbf{m^{a}}$ for a model $m$ is given by:

\begin{equation}
\mathbf{m^{a}} = (m^{a}_{0}, m^{a}_{1}, \ldots, m^{a}_{n-1}, m^{a}_{n})
\end{equation}

\noindent The measurement $\textrm{CAT}_{m}(a,b)$ between attributes $a$ and $b$ in a model $m$ is given by Spearman's $\rho$:

\begin{equation}
    \textrm{CAT}_{m}(a,b) = \rho(\mathbf{m^{a}},\mathbf{m^{b}})
\end{equation}

\subsection{Subspace Projection for Text-to-Image Models}

We draw on subspace projection methods used by \citet{bolukbasi2016man} and  \citet{omrani2023evaluating} to measure facial impression biases in generative text-to-image models. First, we first obtain image embeddings for the 1,004 OMI images from the top layer of a ViT-Large-Patch32-384 model pretrained on ImageNet. For each attribute $a$, we learn a weights vector $w^{a}$ predicting the OMI attribute ratings $h^{a}$, corresponding to a semantic subspace in the embeddings for the attribute. We then use a generative model $g$ to generate $n$ images via a prompt corresponding to either the positive ($a^{+}$) or negative ($a^{-}$) pole of an attribute. We embed each generated image $g_{j}$ at position $j$ with the ViT-Large-Patch32-384 to obtain the vector $\vec{g_{j}}$, and compute an attribute association $g^{a}_{j}$ as its projection product with $w^{a}$:

\begin{equation}
    g^{a}_{j} = \frac{\vec{g_{j}} \cdot w^{a}}{||w^{a}||}
\end{equation}

\noindent The vector of associations $g^{a}$ for a generative model $g$ with an attribute $a$ is given by:

\begin{equation}
    \mathbf{g^{a}} = (g^{a}_{0}, g^{a}_{1}, \ldots, g^{a}_{n-1}, g^{a}_{n})
\end{equation}

\begin{figure*}
\begin{tikzpicture}
\centering
\begin{axis}[
legend pos=outer north east,
width=.97\textwidth,
height=5cm,
ymin=-.55,
xmin=0,
ymajorgrids=true,
ylabel={Model-Human Similarity},
xlabel={Human Inter-Rater Reliability},
xtick style={draw=none},
title={Mean Model-Human Similarity vs. Human IRR for 34 OMI Attributes},
ylabel near ticks,
ylabel shift=-.2,
legend style={
    at={(0.14,.44)}, 
    anchor=south,
    legend image post style={scale=2},
    }
]
\addplot [only marks, mark = *, mark size=2, mark options={color=red!70, draw=black, opacity=0.5, thick}] table[x=Human, y=OpenAI] {\datatable};
\addplot [thick, dashdotted, red] table[
    x=Human,
    y={create col/linear regression={y=OpenAI}}
]
{\datatable};
\addplot [only marks, mark = *, mark size=2, mark options={color=yellow, draw=black, opacity=0.5, thick}] table[x=Human, y=FaceCLIP] {\datatable};

\addplot [very thick, dashed, yellow] table[
    x=Human,
    y={create col/linear regression={y=FaceCLIP}}
]
{\datatable};
\addplot [only marks, mark = *, mark size=2, mark options={color=cyan, draw=black, opacity=0.5, thick}] table[x=Human, y=Scaling] {\datatable};

\addplot [very thick, loosely dotted, blue] table[
    x=Human,
    y={create col/linear regression={y=Scaling}}
]
{\datatable};
\addlegendentry{OpenAI}
\addlegendentry{}
\addlegendentry{FaceCLIP}
\addlegendentry{}
\addlegendentry{Scaling}
\addlegendentry{}
\end{axis}
\end{tikzpicture}
\caption{\small The similarity of CLIP bias to human bias is strongly correlated with human IRR, indicating that the societal consistency of a bias plays a significant role in whether a model learns it during semi-supervised pretraining.}
\label{fig:clip_irr_scatterplot}
\end{figure*}
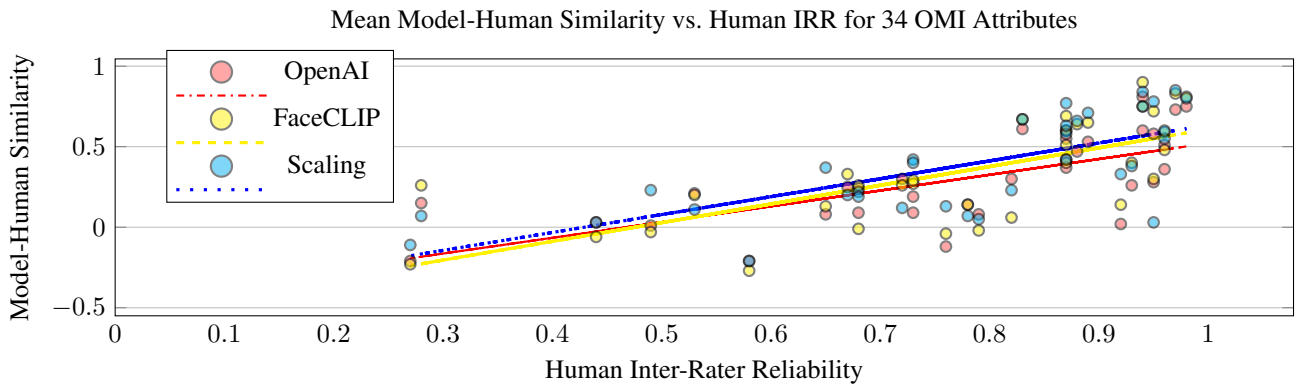

\section{Experiments}

Four experiments test the existence of human-like facial impression bias in vision-language AI, with consideration given to Human IRR, model and dataset scale, and downstream impact in image generation.

\subsection{Model vs. Human Biases}

We tested whether OpenAI, Scaling, and FaceCLIP CLIP models reflect human-like facial impression biases. We obtained the human-model similarity $s^{a}_{m}$ for each model $m$ with each attribute $a$ for the 34 OMI attributes. We then compared the mean human-model similarity for each group of models to the Human IRR for the attribute reported by \citet{peterson2022deep}, calculating Pearson's $\rho$ between Human IRR and model-human similarity for each of the 34 attributes. A large coefficient indicates that the more societally consistent a facial impression bias (\textit{i.e.}, as Human IRR increases), the more likely the bias is to be learned during semi-supervised CLIP training. We also computed Pearson's $\rho$ pairwise between OpenAI, Scaling, and FaRL models to assess whether models trained on different datasets learn similar biases.

\subsection{Effects of Dataset Scale}

We calculated human-model similarity $s^{a}_{m}$ for each model $m$ with each attribute $a$ for the 34 attributes studied, and we constructed a multiple linear regression to predict the model-human similarity $s^{a}_{m}$ for a given CLIP model $m$ and an attribute $a$. We examined the 27 CLIP models trained by \citet{cherti2022reproducible} and produced via the combination of three CLIP architectures (ViT-B32, ViT-B16, and ViT-L14), three dataset sizes (80M, 400M, 2B), and three total training example counts (3B, 13B, 34B). Independent variables include Human IRR (from Peterson et al. \cite{peterson2022deep}), as well as Dataset Size, Model Parameter Count, and Total Training Examples (from \citet{cherti2022reproducible}). We normalized variables with range outside of $(0,1)$ by dividing by their max. We conducted post hoc comparisons between each level of scale (80m, 400m, 2b), using paired-samples $t$-tests.

\subsection{Structure of Facial Impressions}

We computed the correlation matrix $\mathbf{C_m}$ for the 27 CLIP models studied in the Dataset Scale analysis by obtaining $\textrm{CAT}_{m}(a,b)$ for every attribute pair $a \in A$ and $b \in A$. We computed a corresponding matrix $\mathbf{C_h}$ by obtaining correlations for the same attributes using the human ratings of the OMI dataset. We then measured the similarity of $C_m$ and $C_h$ based on the normalized Frobenius inner product $\textrm{F}_{m, h}$. We used a one-way ANOVA to test for differences in $\textrm{F}_{m, h}$ between the 80M, 400M, and 2B models. We used a paired-samples $t$-test to conduct post hoc comparisons.

\citet{peterson2022deep} study the structure of facial impressions by computing the correlation matrix of OMI ratings and qualitatively examining its hierarchical structure. The present research also qualitatively compares the structure of the OMI correlation matrix to the hierarchically clustered correlation matrix of a CLIP-ViT-L-14 trained on LAION-2B for 13-billion total samples. If the structure of OMI  biases is similar to CLIP, we expect to observe similar attribute clusters at higher levels, with differences emerging in leaf nodes.

\subsection{Generative Text-to-Image Models}

Finally, we extended the analysis to SDXL-Turbo, SD2, and Runway SD1.5. We first studied the similarity of these models' representations to human attribute ratings. To do so, we generated $N$=25 images for each attribute's positive pole and 25 for its negative pole. We adjusted prompts for the models to ``a realistic portrait photo of someone who is \textit{attribute},'' because image generators may create cartoonish images that do not center the face. We extracted embeddings for these images and computed their projection products $g^{a_{+}}$ and $g^{a_{-}}$. If generative text-to-image models reflect human-like facial impression biases, we expect images generated from a positive prompt to have positive projections, and images from a negative prompt to have negative projections. We thus frame model-human similarity as a classification problem, wherein images generated from the positive pole prompt receives a label of 1, and from the negative pole prompt receive a label of 0. The OMI subspace is positioned as a classifier, which predicts 1 where an image vector's projection product is positive, and 0 where it is negative. We report Recall, Precision, and F1 Score. We validated this approach using the Outdoors attribute, a control group for measuring validity in the OMI dataset, obtaining F1=.94 for SDXL-Turbo

We then study social bias in Stable Diffusion XL-Turbo by projecting the positive prompt images generated for the White and Black attributes onto all 34 of the OMI attribute subspaces. For each subspace, we compute the White-Black differential bias by obtaining an effect size (Cohen's $d$) between the projection products $\mathbf{g^{White^{+}}}$ and $\mathbf{g^{Black^{+}}}$, and we then measure statistical significance using a paired samples $t$-test.
\begin{table*}[htbp]
\fontsize{7.5}{8}\selectfont
\centering
\begin{tabular}
{|l||r|r|r|c|c|c|S[table-format=3.2]|S[table-format=3.2]|S[table-format=3.2]|}
 \hline
 \multicolumn{10}{|c|}{Model-Human Similarity of Facial Impression Bias in CLIP Models by Pretraining Dataset Size} \\
 \hline
 Measurement & \multicolumn{3}{|c|}{Mean (Std)} & \multicolumn{3}{|c|}{Max} & \multicolumn{3}{|c|}{Cohen's $d$} \\
\hline
Attribute & \multicolumn{1}{|c|}{2b} & \multicolumn{1}{|c|}{400m} & \multicolumn{1}{|c|}{80m} & \multicolumn{1}{|c|}{2b} & \multicolumn{1}{|c|}{400m} & \multicolumn{1}{|c|}{80m} & \multicolumn{1}{|c|}{2b-80m} & \multicolumn{1}{|c|}{400m-80m} & \multicolumn{1}{|c|}{2b-400m} \\
\hline
Happy & \gray{88}.88 (0.02) & \gray{86}.86 (0.03) & \gray{76}.76 (0.06) & \gray{91}.91 & \gray{90}.90 & \gray{84}.84 & \gray{100}1.54* & \gray{100}1.36* &\gray{73}.73  \\
Gender & \gray{85}.85 (0.02) & \gray{87}.87 (0.02) & \gray{83}.83 (0.04) & \gray{87}.87 & \gray{91}.91 & \gray{89}.89 & \gray{61}.61 & \gray{100}1.08* &\gray{0}-.84*  \\
Age & \gray{83}.83 (0.08) & \gray{84}.84 (0.05) & \gray{73}.73 (0.11) & \gray{93}.93 & \gray{90}.90 & \gray{90}.90 & \gray{94}.94 & \gray{100}1.03* &\gray{0}-.04  \\
Asian & \gray{82}.82 (0.03) & \gray{81}.81 (0.05) & \gray{73}.73 (0.06) & \gray{85}.85 & \gray{86}.86 & \gray{79}.79 & \gray{100}1.40* & \gray{100}1.13* &\gray{43}.43  \\
Hispanic & \gray{82}.82 (0.02) & \gray{80}.80 (0.02) & \gray{67}.67 (0.07) & \gray{85}.85 & \gray{84}.84 & \gray{78}.78 & \gray{100}1.62* & \gray{100}1.50* &\gray{100}1.07*  \\
Outdoors & \gray{81}.81 (0.01) & \gray{79}.79 (0.04) & \gray{65}.65 (0.10) & \gray{82}.82 & \gray{84}.84 & \gray{75}.75 & \gray{100}1.51* & \gray{100}1.39* &\gray{57}.57  \\
Pacific Islander & \gray{76}.76 (0.03) & \gray{74}.74 (0.04) & \gray{62}.62 (0.11) & \gray{81}.81 & \gray{81}.81 & \gray{77}.77 & \gray{100}1.28* & \gray{100}1.15* &\gray{50}.50  \\
Middle Eastern & \gray{72}.72 (0.02) & \gray{68}.68 (0.05) & \gray{57}.57 (0.11) & \gray{75}.75 & \gray{78}.78 & \gray{71}.71 & \gray{100}1.43* & \gray{100}1.17* &\gray{88}.88  \\
Native American & \gray{68}.68 (0.06) & \gray{67}.67 (0.06) & \gray{53}.53 (0.13) & \gray{78}.78 & \gray{78}.78 & \gray{77}.77 & \gray{100}1.20* & \gray{100}1.15* &\gray{14}.14  \\
Weight & \gray{68}.68 (0.03) & \gray{69}.69 (0.03) & \gray{63}.63 (0.05) & \gray{72}.72 & \gray{72}.72 & \gray{67}.67 & \gray{100}1.05 & \gray{100}1.24* &\gray{0}-.40  \\
Hair-Color & \gray{66}.66 (0.07) & \gray{61}.61 (0.09) & \gray{50}.50 (0.12) & \gray{76}.76 & \gray{70}.70 & \gray{67}.67 & \gray{100}1.31* & \gray{92}.92 &\gray{67}.67*  \\
Cute & \gray{65}.65 (0.10) & \gray{67}.67 (0.08) & \gray{41}.41 (0.15) & \gray{78}.78 & \gray{76}.76 & \gray{56}.56 & \gray{100}1.35* & \gray{100}1.46* &\gray{0}-.31  \\
Long-Haired & \gray{63}.63 (0.10) & \gray{63}.63 (0.11) & \gray{43}.43 (0.16) & \gray{76}.76 & \gray{75}.75 & \gray{67}.67 & \gray{100}1.20* & \gray{100}1.16* &\gray{6}.06  \\
Gay & \gray{49}.49 (0.07) & \gray{41}.41 (0.06) & \gray{30}.30 (0.11) & \gray{57}.57 & \gray{51}.51 & \gray{42}.42 & \gray{100}1.46* & \gray{100}1.09* &\gray{100}1.14*  \\
Attractive & \gray{48}.48 (0.08) & \gray{50}.50 (0.11) & \gray{28}.28 (0.21) & \gray{60}.60 & \gray{65}.65 & \gray{57}.57 & \gray{100}1.11* & \gray{100}1.11* &\gray{0}-.16  \\
Electable & \gray{47}.47 (0.22) & \gray{50}.50 (0.11) & \gray{30}.30 (0.25) & \gray{68}.68 & \gray{65}.65 & \gray{60}.60 & \gray{72}.72 & \gray{94}.94 &\gray{0}-.16  \\
Smart & \gray{42}.42 (0.15) & \gray{45}.45 (0.14) & \gray{25}.25 (0.11) & \gray{62}.62 & \gray{59}.59 & \gray{50}.50 & \gray{100}1.09* & \gray{100}1.23* &\gray{0}-.20  \\
Black & \gray{41}.41 (0.09) & \gray{35}.35 (0.11) & \gray{37}.37 (0.12) & \gray{51}.51 & \gray{49}.49 & \gray{51}.51 & \gray{46}.46 & \gray{0}-.10 &\gray{59}.59  \\
Smug & \gray{38}.38 (0.11) & \gray{19}.19 (0.20) & \gray{07}.07 (0.15) & \gray{50}.50 & \gray{45}.45 & \gray{25}.25 & \gray{100}1.50* & \gray{63}.63 &\gray{100}1.02*  \\
Trustworthy & \gray{36}.36 (0.18) & \gray{04}.04 (0.18) & \gray{0}-.06 (0.15) & \gray{59}.59 & \gray{35}.35 & \gray{15}.15 & \gray{100}1.56* & \gray{60}.60* &\gray{100}1.33*  \\
Skin-Color & \gray{36}.36 (0.14) & \gray{37}.37 (0.13) & \gray{28}.28 (0.11) & \gray{59}.59 & \gray{58}.58 & \gray{43}.43 & \gray{58}.58 & \gray{70}.70 &\gray{0}-.11  \\
Outgoing & \gray{33}.33 (0.13) & \gray{21}.21 (0.17) & \gray{16}.16 (0.16) & \gray{52}.52 & \gray{35}.35 & \gray{42}.42 & \gray{100}1.02* & \gray{31}.31 &\gray{74}.74  \\
Privileged & \gray{26}.26 (0.15) & \gray{21}.21 (0.12) & \gray{05}.05 (0.07) & \gray{49}.49 & \gray{44}.44 & \gray{16}.16 & \gray{100}1.35* & \gray{100}1.31* &\gray{33}.33  \\
Godly & \gray{20}.20 (0.09) & \gray{26}.26 (0.17) & \gray{25}.25 (0.10) & \gray{34}.34 & \gray{51}.51 & \gray{42}.42 & \gray{0}-.57 & \gray{1}.01 &\gray{0}-.42  \\
Liberal & \gray{14}.14 (0.15) & \gray{22}.22 (0.15) & \gray{23}.23 (0.19) & \gray{32}.32 & \gray{45}.45 & \gray{50}.50 & \gray{0}-.49 & \gray{0}-.05 &\gray{0}-.51  \\
Typical & \gray{13}.13 (0.10) & \gray{11}.11 (0.16) & \gray{05}.05 (0.13) & \gray{28}.28 & \gray{28}.28 & \gray{20}.20 & \gray{72}.72 & \gray{46}.46 &\gray{16}.16  \\
Dorky & \gray{13}.13 (0.23) & \gray{17}.17 (0.22) & \gray{05}.05 (0.21) & \gray{38}.38 & \gray{52}.52 & \gray{33}.33 & \gray{37}.37 & \gray{52}.52 &\gray{0}-.15  \\
Familiar & \gray{12}.12 (0.12) & \gray{04}.04 (0.08) & \gray{0}-.00 (0.14) & \gray{32}.32 & \gray{14}.14 & \gray{16}.16 & \gray{90}.90* & \gray{42}.42 &\gray{75}.75*  \\
Well-Groomed & \gray{12}.12 (0.26) & \gray{0}-.04 (0.21) & \gray{11}.11 (0.18) & \gray{43}.43 & \gray{44}.44 & \gray{31}.31 & \gray{3}.03 & \gray{0}-.71 &\gray{62}.62  \\
Dominant & \gray{11}.11 (0.23) & \gray{06}.06 (0.18) & \gray{00}.00 (0.29) & \gray{47}.47 & \gray{34}.34 & \gray{29}.29 & \gray{41}.41 & \gray{25}.25 &\gray{23}.23  \\
Memorable & \gray{08}.08 (0.11) & \gray{0}-.05 (0.15) & \gray{04}.04 (0.12) & \gray{24}.24 & \gray{13}.13 & \gray{18}.18 & \gray{35}.35 & \gray{0}-.67* &\gray{92}.92*  \\
White & \gray{00}.00 (0.28) & \gray{12}.12 (0.21) & \gray{0}-.03 (0.20) & \gray{45}.45 & \gray{45}.45 & \gray{26}.26 & \gray{12}.12 & \gray{67}.67 &\gray{0}-.46  \\
Looks-Like-You & \gray{0}-.13 (0.14) & \gray{0}-.10 (0.14) & \gray{0}-.11 (0.16) & \gray{15}.15 & \gray{06}.06 & \gray{11}.11 & \gray{0}-.17 & \gray{8}.08 &\gray{0}-.26  \\
Alert & \gray{0}-.29 (0.10) & \gray{0}-.17 (0.10) & \gray{0}-.14 (0.09) & \gray{0}-.16 & \gray{0}-.03 & \gray{05}.05 & \gray{0}-1.25* & \gray{0}-.31 &\gray{0}-1.05*  \\
\hline
 \end{tabular}
 \caption{\small 17 of 34 OMI attributes exhibit statistically significant differences and large effect sizes between model groups.}
 \label{effect_size_table}
 \end{table*}

\section{Results}

Our results indicate that 1) CLIP models exhibit facial impression biases; 2) Human IRR is a significant predictor of which biases are learned; 3) models trained on larger datasets exhibit emergence of subjective facial impression biases, and more human-like associations among impressions; and 4) text-to-image generators exhibit facial impression biases and undesirable social biases associating preferred attributes with images of White individuals.

\begin{figure}[]
\centering
\begin{tikzpicture}
  \def\correlationMatrix{
    {1.00, 0, 0, 0},
    {0.73, 1.00, 0, 0},
    {0.76, 0.94, 1.00, 0},
    {0.72, 0.82, 0.88, 1.00}
  }

  \matrix (m) [
    matrix of nodes,
    nodes={minimum width=1.25cm, minimum height=1.25cm, anchor=center, draw=black},
    row sep=-\pgflinewidth,
    column sep=-\pgflinewidth,
  ] 
  {
    |[fill=purple!100]| 1.00 & & & \\
    |[fill=purple!31]| 0.73 & |[fill=purple!100]| 1.00 & & \\
    |[fill=purple!40]| 0.76 & |[fill=purple!94]| 0.94 & |[fill=purple!100]| 1.00 & \\
    |[fill=purple!28]| 0.72 & |[fill=purple!58]| 0.82 & |[fill=purple!76]| 0.88 & |[fill=purple!100]| 1.00 \\
  };

  \node[below] at (m-4-1.south) {\small OMI IRR};
  \node[below] at (m-4-2.south) {\small OpenAI};
  \node[below] at (m-4-3.south) {\small FaceCLIP};
  \node[below] at (m-4-4.south) {\small Scaling};

  \node[left, rotate=90] at ([xshift=-.25cm, yshift=.5cm]m-4-1.west) {\small Scaling};
  \node[left, rotate=90] at ([xshift=-.25cm, yshift=.75cm]m-3-1.west) {\small FaceCLIP};
  \node[left, rotate=90] at ([xshift=-.25cm, yshift=.65cm]m-2-1.west) {\small OpenAI};
  \node[left, rotate=90] at ([xshift=-.25cm, yshift=.75cm]m-1-1.west) {\small OMI IRR};

  \useasboundingbox (m-1-1.north west) rectangle (m-4-4.south east);

\end{tikzpicture}
\caption{\small CLIP models exhibit significant Spearman's $\rho$ between Mean Model-Human Similarity and OMI IRR.}
\label{correlation_matrix}
\end{figure}
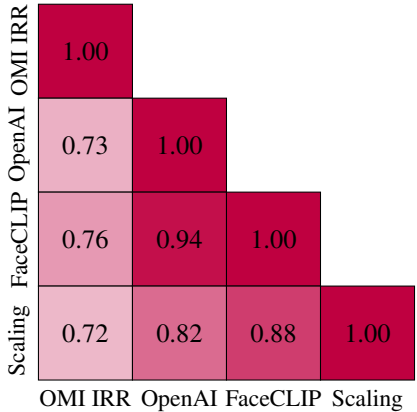

\subsection{CLIP Models Reflect Human Biases}

As shown in Figure \ref{fig:model_human_similarity_bar_chart}, OpenAI, FaceCLIP, and Scaling CLIP models exhibit human-like facial impression biases. Relatively objective attributes like age, hair-color, and happiness exhibit high model-human similarity, as do many socially constructed attributes such as gender and cuteness. Notable exceptions include Black, White, and skin color attributes, which fall short of expectations based on Human IRR, as visualized in Figure \ref{fig:clip_irr_scatterplot}. Model-human similarities of traits like Trustworthiness, Electability, and Intelligence are statistically significant but lower, consistent with the lower IRR of these attributes. That CLIP learns these biases at all is noteworthy: to our knowledge, this is the first research to document unobservable facial impression biases learned by a semi-supervised vision-language model (rather than a supervised model of facial impression biases) that are consistent with human societal biases.

As described in Figure \ref{correlation_matrix}, all three families of models exhibit strong correlations ranging from .72 to .76 between the mean model-human similarity of a trait and its Human IRR. Coefficients are larger between OpenAI and FaceCLIP models than with Scaling CLIP models, likely a result of FaceCLIP post-training from OpenAI base models. \looseness=-1

\begin{figure*}
    \centering
    \includegraphics[width=\textwidth]{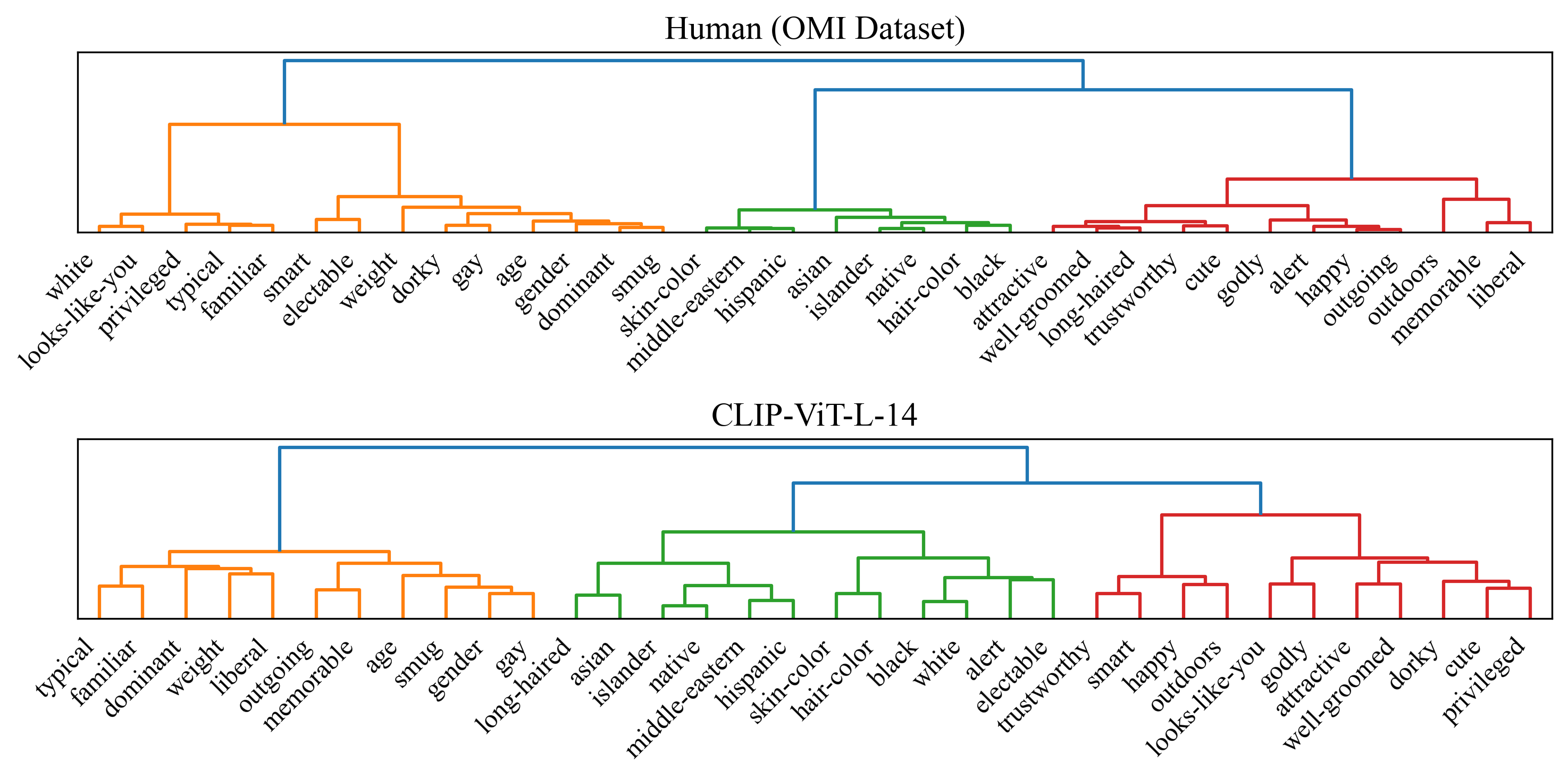}
    \caption{\small The structure of facial impression biases in CLIP-ViT-L-14 mirrors that of human facial impression biases quantified in the OMI dataset. Clusters related to ethnicity emerge in each, as do clusters grouping gender, sexuality, and smugness.}
    \label{fig:dendrogram}
\end{figure*}

\subsection{Dataset Scale}

A multiple linear regression finds that only Human IRR and Dataset Size are statistically significant predictors of model-human similarity. As described in Table \ref{tab:multiple_regression_table}, Human IRR plays the larger role of the two independent variables, as evidenced by a much larger coefficient and $t$-value. Total Training Samples, Image Parameters, and Text Parameters are not statistically significant predictors of facial impression bias in vision-language models.

\begin{table}
\small
\centering 
\begin{tabular}{@{} l *{3}{d{2.6}} @{}} 
\toprule
\multicolumn{1}{l}{Statistic} &  &  &  \\ 
\midrule
    Adj. $R^{2}$ & \multicolumn{1}{c}{.420} & &  \\
    F-Statistic & \multicolumn{1}{c}{134.0} & & \\
   N & \multicolumn{1}{c}{918} & &  \\
   DoF Residuals & \multicolumn{1}{c}{912} &  &  \\
   DoF Model & \multicolumn{1}{c}{5} &  &  \\
\midrule
\multicolumn{1}{l}{Ind. Variable} & \mc{Coef} & \mc{$t$} & \mc{$p<|t|$} \\ 
\midrule
 Human IRR & 1.1204 & 25.470 & .001 \\ 
 Dataset Size & .0877 & 4.484 & .001 \\ 
 Total Samples & -.0132 & -.599 & .549 \\ 
 Image Params & -.5615 & -.150 & .881 \\ 
 Text Params & .7898 & .144 & .885 \\ 
 Constant & -.7542 & -.434 & .665 \\ 
\bottomrule
\end{tabular} 
  \caption{\small Fitting a linear regression to model-human similarity coefficients reveals that Human IRR and Dataset Scale are significant predictors a bias will be learned by a CLIP model.} 
  \label{tab:multiple_regression_table}

\end{table}

Table \ref{effect_size_table} describes the mean (with standard deviation) and maximum model-human similarity for each dataset size, and it reports Cohen’s $d$ between the groups of models trained on the three dataset sizes. Effect sizes obtained between the 400M and the 80M level are large and statistically significant for 17 out of 34 attributes, with large absolute differences between attribute means, such as .41 for Cute at the 80M level vs. .67 at the 400M level. For most attributes, comparisons are not statistically significant between the 2B and 400M levels, though they may return small or medium effect sizes. A notable exception emerges for several unobservable attributes, including Trustworthiness and Sexuality, which exhibit large effect sizes and statistically significant differences between the 2B level and the 400M level. While observable attributes such as Happpiness reflect human ratings well at the 80M level, and more subjective but still visually observable attributes like Cute are reflected consistently at the 400M level, it is not until the 2B level that CLIP models reflect subjective and visually \textit{unobservable} attributes such as Trustworthiness. The results indicate that increases in the scale of the pretraining data have more significant effects for learning subtle societal biases reflecting attributes with \textit{lower} IRR. Models trained on additional data approximate a distribution that more closely reflects the perceptions of society as a whole, learning to use the biased visual heuristics present in the human-authored captions in the pretraining data, even for unobservable attributes.

\subsection{Structure of Facial Impressions}

Results indicate that dataset size impacts the extent to which the \textit{structure} of facial impression bias in CLIP reflects the structure of the facial impression bias in humans. Figure \ref{fig:dendrogram} visualizes the hierarchical similarities between the attribute cross-correlation matrix for CLIP-ViT-L-14 (the most commonly used CLIP model as of this writing) and the OMI attribute cross-correlation matrix. The most salient similarities between the two include a cluster of correlated racial and ethnic identities, such as Hispanic, Middle-Eastern, Native American, and Pacific Islander, as well as a cluster grouping together the Smugness, Gender, Sexuality, and Age attributes. There are also differences between the model and human ratings: while Trustworthy is correlated with Cute in OMI, it is correlated with Smart and Happy in CLIP. Similarities between CLIP attribute correlations and OMI attribute correlations are more evident at the higher levels of the hierarchy, with differences appearing toward the leaf nodes.

\begin{figure}
  \centering
  \begin{tikzpicture}
    \begin{axis}[
      ylabel={\small Frob. Inner Product},
      ymin=0.35,
      ymax=0.625,
      xmin=0.25,
      xmax=3.75,
      ymajorgrids=true,
      xtick style={draw=none},
      ytick style={draw=none},
      width=\linewidth,
      height=4.7cm,
      ylabel near ticks,
      title={CLIP Similarity to OMI Correlation Matrix},
      xtick={1,2,3},
      xticklabels={Scaling-2B, Scaling-400M, Scaling-80M},
      xticklabel style={rotate=10, font=\small, xshift=-.25},
      yticklabel style={font=\small},
      boxplot/draw direction=y,
    ]

    \addplot+[
      boxplot prepared={lower whisker=0.48150892768325865, lower quartile=0.4860705238754213, median=0.5155390673843453, upper quartile=0.5385375209148023, upper whisker=0.5854705450631402}, color=black, fill=red!65,
    ] coordinates {};
    \addplot+[
      boxplot prepared={lower whisker=0.4390684647708985, lower quartile=0.4671417587109023, median=0.4830015862807727, upper quartile=0.5122463241567164, upper whisker=0.520343880849425}, color=black, fill=orange!65,
    ] coordinates {};    
    \addplot+[
      boxplot prepared={lower whisker=0.37583504823693425, lower quartile=0.4007008538603625, median=0.42147224281738355, upper quartile=0.4522357437342802, upper whisker=0.49808194041687304}, color=black, fill=brown!65,
    ] coordinates {};

  \end{axis}
  \end{tikzpicture}
  \caption{\small Scaling-2B CLIP models exhibit the greatest structural similarity to human facial impression biases.}
  \label{fig:boxplot}
\end{figure}
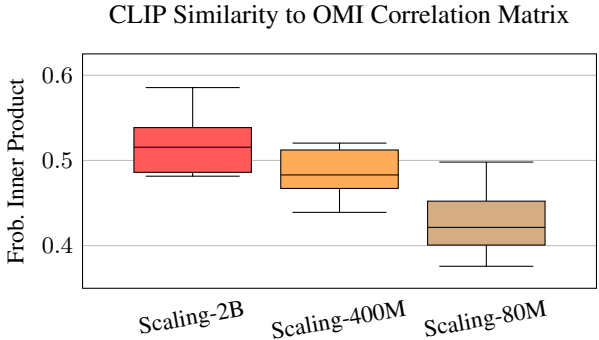

A one-way ANOVA provides evidence that similarities in the structure of facial impression biases increase for models trained on larger datasets, with $F(2) = 15.71$, $p<.001$ after Bonferonni correction, demonstrating statistically significant differences in $\textrm{F}_{m, h}$ between the Scaling-2B, Scaling-400M, and Scaling-80M models. We observe statistically significant differences and large effect sizes both between the 80M and 400M levels and between the 80M and 2B levels with post-hoc tests. Figure \ref{fig:boxplot} illustrates the differences among the three model groups, showing that the magnitude of difference is greater between the 80M level and the 400M level than between the 400M level and the 2B level.

\subsection{Generative Text-to-Image Models}

As shown in Table \ref{tab:texttoimage}, we observe more variance in text-to-image generator F1 scores than in CLIP model-human similarities. SDXL has the most human-like associations, with Spearman's $\rho=.60, p<.001$ between Human IRR and SDXL F1 scores. Runway-SD1.5 is also correlated with IRR, with $\rho=.50, p<.01$, while SD2 is not significantly correlated with IRR, with $\rho=.32, p=.06$. Notably, the Attractive attribute has the highest F1 score for SDXL, whereas it is the 16th most human-similar attribute in CLIP models, suggesting the importance of representing beauty for user-facing image generators, which often undergo additional training to better reflect user aesthetic preferences \cite{rombach2022high}. With the exception of Liberal, unobservable traits rank in the bottom half of F1 scores for SDXL, and Trustworthy is lowest of any trait. Though SD2 and Runway-SD1.5 are more human-like than SDXL for trustworthiness, the results suggest that exploiting biased heuristics may be more straightforward for a classifier like CLIP.

\begin{table}[htbp]
\fontsize{7.5}{8}\selectfont
\centering
\begin{tabular}
{|l||c|c|c|}
 \hline
\multicolumn{4}{|c|}{Attribute F1 Scores by Stable Diffusion Model} \\
 \hline
 Attribute & SDXL & SD2 & Runway-SD1.5 \\
 \hline
attractive & \gray{98}0.98 & \gray{70}0.7 & \gray{65}0.65 \\ 
outdoors & \gray{94}0.94 & \gray{67}0.67 & \gray{60}0.60 \\ 
well-groomed & \gray{91}0.91 & \gray{66}0.66 & \gray{59}0.59 \\ 
hair-color & \gray{83}0.83 & \gray{79}0.79 & \gray{77}0.77 \\ 
weight & \gray{76}0.76 & \gray{56}0.57 & \gray{37}0.37 \\ 
long-haired & \gray{76}0.76 & \gray{67}0.67 & \gray{65}0.65 \\ 
black & \gray{75}0.75 & \gray{76}0.76 & \gray{76}0.76 \\ 
white & \gray{74}0.74 & \gray{43}0.43 & \gray{53}0.53 \\ 
asian & \gray{70}0.70 & \gray{63}0.63 & \gray{69}0.69 \\ 
middle-eastern & \gray{69}0.69 & \gray{70}0.70 & \gray{64}0.64 \\ 
cute & \gray{69}0.69 & \gray{57}0.58 & \gray{55}0.55 \\ 
happy & \gray{68}0.68 & \gray{79}0.79 & \gray{77}0.77 \\ 
islander & \gray{68}0.68 & \gray{66}0.66 & \gray{69}0.69 \\ 
age & \gray{68}0.68 & \gray{68}0.68 & \gray{72}0.72 \\ 
liberal & \gray{68}0.68 & \gray{60}0.60 & \gray{62}0.62 \\ 
skin-color & \gray{67}0.67 & \gray{68}0.68 & \gray{61}0.61 \\ 
alert & \gray{67}0.67 & \gray{56}0.57 & \gray{59}0.59 \\ 
gender & \gray{66}0.66 & \gray{64}0.64 & \gray{62}0.62 \\ 
smart & \gray{65}0.65 & \gray{56}0.57 & \gray{71}0.71 \\ 
dominant & \gray{65}0.65 & \gray{68}0.68 & \gray{67}0.67 \\ 
hispanic & \gray{65}0.65 & \gray{66}0.66 & \gray{69}0.69 \\ 
native & \gray{64}0.64 & \gray{69}0.69 & \gray{68}0.68 \\ 
electable & \gray{61}0.61 & \gray{65}0.65 & \gray{60}0.60 \\ 
dorky & \gray{53}0.53 & \gray{60}0.60 & \gray{60}0.60 \\ 
looks-like-you & \gray{53}0.53 & \gray{68}0.68 & \gray{49}0.49 \\ 
smug & \gray{51}0.51 & \gray{57}0.58 & \gray{56}0.57 \\ 
memorable & \gray{48}0.48 & \gray{66}0.66 & \gray{59}0.59 \\ 
privileged & \gray{47}0.47 & \gray{60}0.60 & \gray{57}0.58 \\ 
gay & \gray{37}0.37 & \gray{56}0.56 & \gray{40}0.40 \\ 
godly & \gray{31}0.31 & \gray{65}0.65 & \gray{59}0.59 \\ 
typical & \gray{28}0.29 & \gray{63}0.63 & \gray{63}0.63 \\ 
familiar & \gray{27}0.27 & \gray{68}0.68 & \gray{51}0.51 \\ 
outgoing & \gray{24}0.24 & \gray{66}0.66 & \gray{71}0.71 \\ 
trustworthy & \gray{12}0.12 & \gray{51}0.51 & \gray{64}0.64 \\ 
\hline
\end{tabular}
\caption{\small SDXL-Turbo reflects human facial impression biases, with Spearman's $\rho=.60$ between IRR and SDXL F1.}
\label{tab:texttoimage}
\end{table}

\begin{figure*}
  \centering
  \begin{tikzpicture}
    \begin{axis}[
      ylabel={\small Projection Products},
        ylabel style={rotate=-90, at={(0,0.5)}},
        ylabel near ticks,
      ymin=-15,
      ymax=15,
      xmin=0,
      xmax=69,
      xtick style={draw=none},
      ytick style={draw=none},
      width=\textwidth,
      height=5cm,
      xlabel near ticks,
      title={Differential White-Black Bias in Stable-Diffusion-XL-Turbo},
      xtick={1.7, 2.7, 3.7, 4.7, 5.7, 6.7, 7.7, 8.7, 9.7, 10.7, 11.7, 12.7, 13.7, 14.7, 15.7, 16.7, 17.7, 18.7, 19.7, 20.7, 21.7, 22.7, 23.7, 24.7, 25.7, 26.7, 27.7, 28.7, 29.7, 30.7, 31.7, 32.7, 33.7, 34.7, 35.7, 36.7, 37.7, 38.7, 39.7, 40.7, 41.7, 42.7, 43.7, 44.7, 45.7, 46.7, 47.7, 48.7, 49.7, 50.7, 51.7, 52.7, 53.7, 54.7, 55.7, 56.7, 57.7, 58.7, 59.7, 60.7, 61.7, 62.7, 63.7, 64.7, 65.7, 66.7, 67.7},
      xticklabels={dominant, , white, , privileged, , memorable, , dorky, , alert, , age, , electable, , outdoors, , attractive, , long-haired, , happy, , well-groomed, , typical, , smart, , cute, , gender, , outgoing, , smug, , familiar, , looks-like-you, , middle-eastern, , trustworthy, , asian, , native, , hispanic, , liberal, , godly, , gay, , weight, , islander, , hair-color, , skin-color, , black},
      ytick={-12,-10.5,-5,0,5,10},
      yticklabels={{\tiny \textit{d}}, {\tiny \textit{p}}, -5, 0, 5, 10},
      yticklabel style={font=\small, yshift=-.5},
      xticklabel style={font=\small, xshift=-.25,style={rotate=-90}},
      boxplot/draw direction=y
    ]
\addplot+[boxplot prepared={lower whisker=5.358373260319571, lower quartile=5.358373260319571, median=6.4432695251330285, upper quartile=8.181930666311429, upper whisker=8.181930666311429}, color=black, fill=blue!65, solid] coordinates {};
\addplot+[boxplot prepared={lower whisker=-1.891667091940376, lower quartile=-1.891667091940376, median=-0.9923926738811514, upper quartile=-0.19755725581658062, upper whisker=-0.19755725581658062}, color=black, fill=red!65, solid] coordinates {};
\addplot+[boxplot prepared={lower whisker=1.7816463846949362, lower quartile=1.7816463846949362, median=4.559221434220414, upper quartile=5.731721138650141, upper whisker=5.731721138650141}, color=black, fill=blue!65, solid] coordinates {};
\addplot+[boxplot prepared={lower whisker=-6.748338438115567, lower quartile=-6.748338438115567, median=-4.902434238040563, upper quartile=-3.2718537289031535, upper whisker=-3.2718537289031535}, color=black, fill=red!65, solid] coordinates {};
\addplot+[boxplot prepared={lower whisker=5.0918820172860135, lower quartile=5.0918820172860135, median=8.64871854575313, upper quartile=9.659749931836922, upper whisker=9.659749931836922}, color=black, fill=blue!65, solid] coordinates {};
\addplot+[boxplot prepared={lower whisker=-6.335483725288996, lower quartile=-6.335483725288996, median=-2.1910094413546712, upper quartile=0.22540931344404425, upper whisker=0.22540931344404425}, color=black, fill=red!65, solid] coordinates {};
\addplot+[boxplot prepared={lower whisker=2.156673146213467, lower quartile=2.156673146213467, median=3.4785833121545195, upper quartile=4.351923755324366, upper whisker=4.351923755324366}, color=black, fill=blue!65, solid] coordinates {};
\addplot+[boxplot prepared={lower whisker=-5.798184277290055, lower quartile=-5.798184277290055, median=-3.872636603484133, upper quartile=-2.008212066766928, upper whisker=-2.008212066766928}, color=black, fill=red!65, solid] coordinates {};
\addplot+[boxplot prepared={lower whisker=2.7934728404013094, lower quartile=2.7934728404013094, median=6.570937673192487, upper quartile=10.99078218581723, upper whisker=10.99078218581723}, color=black, fill=blue!65, solid] coordinates {};
\addplot+[boxplot prepared={lower whisker=-0.823225000319227, lower quartile=-0.823225000319227, median=0.846670975292617, upper quartile=2.638812713378539, upper whisker=2.638812713378539}, color=black, fill=red!65, solid] coordinates {};
\addplot+[boxplot prepared={lower whisker=6.183585612932127, lower quartile=6.183585612932127, median=10.221386268560789, upper quartile=12.49665069029145, upper whisker=12.49665069029145}, color=black, fill=blue!65, solid] coordinates {};
\addplot+[boxplot prepared={lower whisker=0.9154813076194853, lower quartile=0.9154813076194853, median=2.683171767175269, upper quartile=6.1318662378875795, upper whisker=6.1318662378875795}, color=black, fill=red!65, solid] coordinates {};
\addplot+[boxplot prepared={lower whisker=6.791718266264667, lower quartile=6.791718266264667, median=10.366041628135122, upper quartile=11.664703080289097, upper whisker=11.664703080289097}, color=black, fill=blue!65, solid] coordinates {};
\addplot+[boxplot prepared={lower whisker=3.802985894895448, lower quartile=3.802985894895448, median=5.5699722481588685, upper quartile=6.360200831229935, upper whisker=6.360200831229935}, color=black, fill=red!65, solid] coordinates {};
\addplot+[boxplot prepared={lower whisker=4.94989041244958, lower quartile=4.94989041244958, median=6.262072650372205, upper quartile=7.580814797484939, upper whisker=7.580814797484939}, color=black, fill=blue!65, solid] coordinates {};
\addplot+[boxplot prepared={lower whisker=0.9076681652888312, lower quartile=0.9076681652888312, median=4.078416578076513, upper quartile=5.018271081580592, upper whisker=5.018271081580592}, color=black, fill=red!65, solid] coordinates {};
\addplot+[boxplot prepared={lower whisker=-1.2902561430058628, lower quartile=-1.2902561430058628, median=0.8520003298629015, upper quartile=2.424971640947957, upper whisker=2.424971640947957}, color=black, fill=blue!65, solid] coordinates {};
\addplot+[boxplot prepared={lower whisker=-5.466272451244167, lower quartile=-5.466272451244167, median=-3.375855583289735, upper quartile=-2.161009616992038, upper whisker=-2.161009616992038}, color=black, fill=red!65, solid] coordinates {};
\addplot+[boxplot prepared={lower whisker=0.6376144637734679, lower quartile=0.6376144637734679, median=2.5773477482115723, upper quartile=3.9610948342265524, upper whisker=3.9610948342265524}, color=black, fill=blue!65, solid] coordinates {};
\addplot+[boxplot prepared={lower whisker=-3.489670936819916, lower quartile=-3.489670936819916, median=-0.6931796852667881, upper quartile=0.31226297062505143, upper whisker=0.31226297062505143}, color=black, fill=red!65, solid] coordinates {};
\addplot+[boxplot prepared={lower whisker=-2.6014700076712467, lower quartile=-2.6014700076712467, median=0.3147238743976725, upper quartile=2.168191913725693, upper whisker=2.168191913725693}, color=black, fill=blue!65, solid] coordinates {};
\addplot+[boxplot prepared={lower whisker=-6.596811203934483, lower quartile=-6.596811203934483, median=-2.812230175210563, upper quartile=-0.056911869613396684, upper whisker=-0.056911869613396684}, color=black, fill=red!65, solid] coordinates {};
\addplot+[boxplot prepared={lower whisker=1.1412920393189814, lower quartile=1.1412920393189814, median=3.9658292003220104, upper quartile=5.851841102355193, upper whisker=5.851841102355193}, color=black, fill=blue!65, solid] coordinates {};
\addplot+[boxplot prepared={lower whisker=-0.37155474834103647, lower quartile=-0.37155474834103647, median=0.3918242746552058, upper quartile=2.1912974382346424, upper whisker=2.1912974382346424}, color=black, fill=red!65, solid] coordinates {};
\addplot+[boxplot prepared={lower whisker=3.307923621545839, lower quartile=3.307923621545839, median=4.312382737613941, upper quartile=6.591815487209684, upper whisker=6.591815487209684}, color=black, fill=blue!65, solid] coordinates {};
\addplot+[boxplot prepared={lower whisker=-1.8325703482770601, lower quartile=-1.8325703482770601, median=1.5468550394458813, upper quartile=4.277673077364528, upper whisker=4.277673077364528}, color=black, fill=red!65, solid] coordinates {};
\addplot+[boxplot prepared={lower whisker=-0.9520564964192629, lower quartile=-0.9520564964192629, median=0.4440624605949593, upper quartile=3.526169914990347, upper whisker=3.526169914990347}, color=black, fill=blue!65, solid] coordinates {};
\addplot+[boxplot prepared={lower whisker=-2.1316050126565047, lower quartile=-2.1316050126565047, median=-0.35959942702524705, upper quartile=2.2752884051945323, upper whisker=2.2752884051945323}, color=black, fill=red!65, solid] coordinates {};
\addplot+[boxplot prepared={lower whisker=4.808867938548471, lower quartile=4.808867938548471, median=6.981543815536172, upper quartile=8.603288271975655, upper whisker=8.603288271975655}, color=black, fill=blue!65, solid] coordinates {};
\addplot+[boxplot prepared={lower whisker=2.50117050743458, lower quartile=2.50117050743458, median=4.353227707676888, upper quartile=6.345465157081875, upper whisker=6.345465157081875}, color=black, fill=red!65, solid] coordinates {};
\addplot+[boxplot prepared={lower whisker=-2.4514629649913795, lower quartile=-2.4514629649913795, median=0.11029849522417452, upper quartile=2.3047151213966783, upper whisker=2.3047151213966783}, color=black, fill=blue!65, solid] coordinates {};
\addplot+[boxplot prepared={lower whisker=-4.756011713857045, lower quartile=-4.756011713857045, median=-1.3981988826426668, upper quartile=0.686208844440281, upper whisker=0.686208844440281}, color=black, fill=red!65, solid] coordinates {};
\addplot+[boxplot prepared={lower whisker=2.330637609213321, lower quartile=2.330637609213321, median=4.789668430632132, upper quartile=6.989523244232733, upper whisker=6.989523244232733}, color=black, fill=blue!65, solid] coordinates {};
\addplot+[boxplot prepared={lower whisker=2.1454819456372385, lower quartile=2.1454819456372385, median=3.7078591655928888, upper quartile=4.824366038196789, upper whisker=4.824366038196789}, color=black, fill=red!65, solid] coordinates {};
\addplot+[boxplot prepared={lower whisker=-0.6736903663588824, lower quartile=-0.6736903663588824, median=5.542631385439472, upper quartile=7.717910534405775, upper whisker=7.717910534405775}, color=black, fill=blue!65, solid] coordinates {};
\addplot+[boxplot prepared={lower whisker=1.1322052556039628, lower quartile=1.1322052556039628, median=2.1145654320030522, upper quartile=3.9446295048656603, upper whisker=3.9446295048656603}, color=black, fill=red!65, solid] coordinates {};
\addplot+[boxplot prepared={lower whisker=-1.1025379996031412, lower quartile=-1.1025379996031412, median=0.3603169920395093, upper quartile=2.0976000593559054, upper whisker=2.0976000593559054}, color=black, fill=blue!65, solid] coordinates {};
\addplot+[boxplot prepared={lower whisker=-2.9660954625665332, lower quartile=-2.9660954625665332, median=-1.485355385381351, upper quartile=0.7940091844392115, upper whisker=0.7940091844392115}, color=black, fill=red!65, solid] coordinates {};
\addplot+[boxplot prepared={lower whisker=-1.915477310117043, lower quartile=-1.915477310117043, median=0.8445584199901011, upper quartile=2.5666651909528024, upper whisker=2.5666651909528024}, color=black, fill=blue!65, solid] coordinates {};
\addplot+[boxplot prepared={lower whisker=-2.104630254950867, lower quartile=-2.104630254950867, median=0.20664006565125803, upper quartile=1.510541041376077, upper whisker=1.510541041376077}, color=black, fill=red!65, solid] coordinates {};
\addplot+[boxplot prepared={lower whisker=-3.036446670968925, lower quartile=-3.036446670968925, median=-1.7084326476858551, upper quartile=0.13434004223661658, upper whisker=0.13434004223661658}, color=black, fill=blue!65, solid] coordinates {};
\addplot+[boxplot prepared={lower whisker=-2.0288383661157843, lower quartile=-2.0288383661157843, median=-0.5032644451742431, upper quartile=0.21399103097919875, upper whisker=0.21399103097919875}, color=black, fill=red!65, solid] coordinates {};
\addplot+[boxplot prepared={lower whisker=-0.295644530159177, lower quartile=-0.295644530159177, median=0.7885333124387293, upper quartile=2.967450503812153, upper whisker=2.967450503812153}, color=black, fill=blue!65, solid] coordinates {};
\addplot+[boxplot prepared={lower whisker=0.4414216860655792, lower quartile=0.4414216860655792, median=2.3009630579473814, upper quartile=4.685312343479039, upper whisker=4.685312343479039}, color=black, fill=red!65, solid] coordinates {};
\addplot+[boxplot prepared={lower whisker=-0.2627248318572683, lower quartile=-0.2627248318572683, median=3.4168503652689193, upper quartile=5.521530396580822, upper whisker=5.521530396580822}, color=black, fill=blue!65, solid] coordinates {};
\addplot+[boxplot prepared={lower whisker=3.0110687733757295, lower quartile=3.0110687733757295, median=4.372849238987219, upper quartile=5.288582508733474, upper whisker=5.288582508733474}, color=black, fill=red!65, solid] coordinates {};
\addplot+[boxplot prepared={lower whisker=-0.05747146205524937, lower quartile=-0.05747146205524937, median=1.8945690446266856, upper quartile=3.813413061550698, upper whisker=3.813413061550698}, color=black, fill=blue!65, solid] coordinates {};
\addplot+[boxplot prepared={lower whisker=-0.0250250421794064, lower quartile=-0.0250250421794064, median=3.0422310848446243, upper quartile=6.881547932236102, upper whisker=6.881547932236102}, color=black, fill=red!65, solid] coordinates {};
\addplot+[boxplot prepared={lower whisker=-7.26937513685628, lower quartile=-7.26937513685628, median=-2.25199983219669, upper quartile=1.2252870628534653, upper whisker=1.2252870628534653}, color=black, fill=blue!65, solid] coordinates {};
\addplot+[boxplot prepared={lower whisker=-2.273698768297811, lower quartile=-2.273698768297811, median=-0.6438176549793877, upper quartile=0.5897610374719771, upper whisker=0.5897610374719771}, color=black, fill=red!65, solid] coordinates {};
\addplot+[boxplot prepared={lower whisker=-3.895740105961988, lower quartile=-3.895740105961988, median=-1.8113746865305518, upper quartile=1.6071569429206682, upper whisker=1.6071569429206682}, color=black, fill=blue!65, solid] coordinates {};
\addplot+[boxplot prepared={lower whisker=0.40769890823004995, lower quartile=0.40769890823004995, median=1.3364898782561259, upper quartile=2.4470480369162497, upper whisker=2.4470480369162497}, color=black, fill=red!65, solid] coordinates {};
\addplot+[boxplot prepared={lower whisker=2.060150333928238, lower quartile=2.060150333928238, median=4.836968042587387, upper quartile=7.005007669738842, upper whisker=7.005007669738842}, color=black, fill=blue!65, solid] coordinates {};
\addplot+[boxplot prepared={lower whisker=4.667581379038131, lower quartile=4.667581379038131, median=6.146098398221164, upper quartile=8.875671575176986, upper whisker=8.875671575176986}, color=black, fill=red!65, solid] coordinates {};
\addplot+[boxplot prepared={lower whisker=-1.5195633285488215, lower quartile=-1.5195633285488215, median=0.6123907720914123, upper quartile=3.1941774735869215, upper whisker=3.1941774735869215}, color=black, fill=blue!65, solid] coordinates {};
\addplot+[boxplot prepared={lower whisker=1.7126225637594266, lower quartile=1.7126225637594266, median=2.5857716924940672, upper quartile=4.583659423523741, upper whisker=4.583659423523741}, color=black, fill=red!65, solid] coordinates {};
\addplot+[boxplot prepared={lower whisker=1.7357335896042014, lower quartile=1.7357335896042014, median=3.5340604498355295, upper quartile=4.405112297618595, upper whisker=4.405112297618595}, color=black, fill=blue!65, solid] coordinates {};
\addplot+[boxplot prepared={lower whisker=3.2841323110963745, lower quartile=3.2841323110963745, median=5.851405344095672, upper quartile=7.2714848448002325, upper whisker=7.2714848448002325}, color=black, fill=red!65, solid] coordinates {};
\addplot+[boxplot prepared={lower whisker=-1.9920840157056476, lower quartile=-1.9920840157056476, median=-0.13825802838742535, upper quartile=1.8250831125834692, upper whisker=1.8250831125834692}, color=black, fill=blue!65, solid] coordinates {};
\addplot+[boxplot prepared={lower whisker=5.507066120198044, lower quartile=5.507066120198044, median=6.522380010007781, upper quartile=8.622010800405306, upper whisker=8.622010800405306}, color=black, fill=red!65, solid] coordinates {};
\addplot+[boxplot prepared={lower whisker=-2.4370213868546076, lower quartile=-2.4370213868546076, median=-0.26357236072995405, upper quartile=2.8755862622171398, upper whisker=2.8755862622171398}, color=black, fill=blue!65, solid] coordinates {};
\addplot+[boxplot prepared={lower whisker=3.9415835998866187, lower quartile=3.9415835998866187, median=5.70267077786898, upper quartile=6.904368050483151, upper whisker=6.904368050483151}, color=black, fill=red!65, solid] coordinates {};
\addplot+[boxplot prepared={lower whisker=-4.944919388990212, lower quartile=-4.944919388990212, median=-2.393393333397811, upper quartile=0.38564818709745197, upper whisker=0.38564818709745197}, color=black, fill=blue!65, solid] coordinates {};
\addplot+[boxplot prepared={lower whisker=2.6718665925553027, lower quartile=2.6718665925553027, median=3.9357725741211578, upper quartile=6.059977629550726, upper whisker=6.059977629550726}, color=black, fill=red!65, solid] coordinates {};
\addplot+[boxplot prepared={lower whisker=1.755185466329182, lower quartile=1.755185466329182, median=2.7135600672653526, upper quartile=3.96289775860073, upper whisker=3.96289775860073}, color=black, fill=blue!65, solid] coordinates {};
\addplot+[boxplot prepared={lower whisker=7.856518448838564, lower quartile=7.856518448838564, median=9.47139843096926, upper quartile=10.09238571454548, upper whisker=10.09238571454548}, color=black, fill=red!65, solid] coordinates {};
\addplot+[boxplot prepared={lower whisker=-1.7702712193199577, lower quartile=-1.7702712193199577, median=-0.6373477546320045, upper quartile=1.3672012343948845, upper whisker=1.3672012343948845}, color=black, fill=blue!65, solid] coordinates {};
\addplot+[boxplot prepared={lower whisker=6.120392466024429, lower quartile=6.120392466024429, median=7.820364695314265, upper quartile=8.812599397085188, upper whisker=8.812599397085188}, color=black, fill=red!65, solid] coordinates {};

    \draw [thick] (axis cs:1,-10) -- (axis cs:1,5.358373260319571);
    \draw [thick] (axis cs:2,-10) -- (axis cs:2,-1.891667091940376);
    \draw [thick] (axis cs:1,-10) -- (axis cs:2,-10,);
    \node at (axis cs:1.6,-11) {\tiny ***};
    \node at (axis cs:1.6,-12) {\tiny 1.75};

    \draw [thick] (axis cs:3,-10) -- (axis cs:3,1.7816463846949362);
    \draw [thick] (axis cs:4,-10) -- (axis cs:4,-6.748338438115567);
    \draw [thick] (axis cs:3,-10) -- (axis cs:4,-10,);
    \node at (axis cs:3.6,-11) {\tiny ***};
    \node at (axis cs:3.6,-12) {\tiny 1.57};

    \draw [thick] (axis cs:5,-10) -- (axis cs:5,5.0918820172860135);
    \draw [thick] (axis cs:6,-10) -- (axis cs:6,-6.335483725288996);
    \draw [thick] (axis cs:5,-10) -- (axis cs:6,-10,);
    \node at (axis cs:5.6,-11) {\tiny ***};
    \node at (axis cs:5.6,-12) {\tiny 1.54};

    \draw [thick] (axis cs:7,-10) -- (axis cs:7,2.156673146213467);
    \draw [thick] (axis cs:8,-10) -- (axis cs:8,-5.798184277290055);
    \draw [thick] (axis cs:7,-10) -- (axis cs:8,-10,);
    \node at (axis cs:7.6,-11) {\tiny ***};
    \node at (axis cs:7.6,-12) {\tiny 1.43};

    \draw [thick] (axis cs:9,-10) -- (axis cs:9,2.7934728404013094);
    \draw [thick] (axis cs:10,-10) -- (axis cs:10,-0.823225000319227);
    \draw [thick] (axis cs:9,-10) -- (axis cs:10,-10,);
    \node at (axis cs:9.6,-11) {\tiny ***};
    \node at (axis cs:9.6,-12) {\tiny 1.26};

    \draw [thick] (axis cs:11,-10) -- (axis cs:11,6.183585612932127);
    \draw [thick] (axis cs:12,-10) -- (axis cs:12,0.9154813076194853);
    \draw [thick] (axis cs:11,-10) -- (axis cs:12,-10,);
    \node at (axis cs:11.6,-11) {\tiny ***};
    \node at (axis cs:11.6,-12) {\tiny 1.20};

    \draw [thick] (axis cs:13,-10) -- (axis cs:13,6.791718266264667);
    \draw [thick] (axis cs:14,-10) -- (axis cs:14,3.802985894895448);
    \draw [thick] (axis cs:13,-10) -- (axis cs:14,-10,);
    \node at (axis cs:13.6,-11) {\tiny ***};
    \node at (axis cs:13.6,-12) {\tiny 1.12};

    \draw [thick] (axis cs:15,-10) -- (axis cs:15,4.94989041244958);
    \draw [thick] (axis cs:16,-10) -- (axis cs:16,0.9076681652888312);
    \draw [thick] (axis cs:15,-10) -- (axis cs:16,-10,);
    \node at (axis cs:15.6,-11) {\tiny ***};
    \node at (axis cs:15.6,-12) {\tiny 1.00};

    \draw [thick] (axis cs:17,-10) -- (axis cs:17,-1.2902561430058628);
    \draw [thick] (axis cs:18,-10) -- (axis cs:18,-5.466272451244167);
    \draw [thick] (axis cs:17,-10) -- (axis cs:18,-10,);
    \node at (axis cs:17.6,-11) {\tiny ***};
    \node at (axis cs:17.6,-12) {\tiny .98};

    \draw [thick] (axis cs:19,-10) -- (axis cs:19,0.6376144637734679);
    \draw [thick] (axis cs:20,-10) -- (axis cs:20,-3.489670936819916);
    \draw [thick] (axis cs:19,-10) -- (axis cs:20,-10,);
    \node at (axis cs:19.6,-11) {\tiny **};
    \node at (axis cs:19.6,-12) {\tiny .95};

    \draw [thick] (axis cs:21,-10) -- (axis cs:21,-2.6014700076712467);
    \draw [thick] (axis cs:22,-10) -- (axis cs:22,-6.596811203934483);
    \draw [thick] (axis cs:21,-10) -- (axis cs:22,-10,);
    \node at (axis cs:21.6,-11) {\tiny **};
    \node at (axis cs:21.6,-12) {\tiny .84};

    \draw [thick] (axis cs:23,-10) -- (axis cs:23,1.1412920393189814);
    \draw [thick] (axis cs:24,-10) -- (axis cs:24,-0.37155474834103647);
    \draw [thick] (axis cs:23,-10) -- (axis cs:24,-10,);
    \node at (axis cs:23.6,-11) {\tiny **};
    \node at (axis cs:23.6,-12) {\tiny .72};

    \draw [thick] (axis cs:25,-10) -- (axis cs:25,3.307923621545839);
    \draw [thick] (axis cs:26,-10) -- (axis cs:26,-1.8325703482770601);
    \draw [thick] (axis cs:25,-10) -- (axis cs:26,-10,);
    \node at (axis cs:25.6,-11) {\tiny *};
    \node at (axis cs:25.6,-12) {\tiny .72};

    \draw [thick] (axis cs:27,-10) -- (axis cs:27,-0.9520564964192629);
    \draw [thick] (axis cs:28,-10) -- (axis cs:28,-2.1316050126565047);
    \draw [thick] (axis cs:27,-10) -- (axis cs:28,-10,);
    \node at (axis cs:27.6,-11) {\tiny \textit{n.s.}};

    \draw [thick] (axis cs:29,-10) -- (axis cs:29,4.808867938548471);
    \draw [thick] (axis cs:30,-10) -- (axis cs:30,2.50117050743458);
    \draw [thick] (axis cs:29,-10) -- (axis cs:30,-10,);
    \node at (axis cs:29.6,-11) {\tiny \textit{n.s.}};

    \draw [thick] (axis cs:31,-10) -- (axis cs:31,-2.4514629649913795);
    \draw [thick] (axis cs:32,-10) -- (axis cs:32,-4.756011713857045);
    \draw [thick] (axis cs:31,-10) -- (axis cs:32,-10,);
    \node at (axis cs:31.6,-11) {\tiny \textit{n.s.}};

    \draw [thick] (axis cs:33,-10) -- (axis cs:33,2.330637609213321);
    \draw [thick] (axis cs:34,-10) -- (axis cs:34,2.1454819456372385);
    \draw [thick] (axis cs:33,-10) -- (axis cs:34,-10,);
    \node at (axis cs:33.6,-11) {\tiny \textit{n.s.}};

    \draw [thick] (axis cs:35,-10) -- (axis cs:35,-0.6736903663588824);
    \draw [thick] (axis cs:36,-10) -- (axis cs:36,1.1322052556039628);
    \draw [thick] (axis cs:35,-10) -- (axis cs:36,-10,);
    \node at (axis cs:35.6,-11) {\tiny \textit{n.s.}};

    \draw [thick] (axis cs:37,-10) -- (axis cs:37,-1.1025379996031412);
    \draw [thick] (axis cs:38,-10) -- (axis cs:38,-2.9660954625665332);
    \draw [thick] (axis cs:37,-10) -- (axis cs:38,-10,);
    \node at (axis cs:37.6,-11) {\tiny \textit{n.s.}};

    \draw [thick] (axis cs:39,-10) -- (axis cs:39,-1.915477310117043);
    \draw [thick] (axis cs:40,-10) -- (axis cs:40,-2.104630254950867);
    \draw [thick] (axis cs:39,-10) -- (axis cs:40,-10,);
    \node at (axis cs:39.6,-11) {\tiny \textit{n.s.}};

    \draw [thick] (axis cs:41,-10) -- (axis cs:41,-3.036446670968925);
    \draw [thick] (axis cs:42,-10) -- (axis cs:42,-2.0288383661157843);
    \draw [thick] (axis cs:41,-10) -- (axis cs:42,-10,);
    \node at (axis cs:41.6,-11) {\tiny \textit{n.s.}};

    \draw [thick] (axis cs:43,-10) -- (axis cs:43,-0.295644530159177);
    \draw [thick] (axis cs:44,-10) -- (axis cs:44,0.4414216860655792);
    \draw [thick] (axis cs:43,-10) -- (axis cs:44,-10,);
    \node at (axis cs:43.6,-11) {\tiny \textit{n.s.}};

    \draw [thick] (axis cs:45,-10) -- (axis cs:45,-0.2627248318572683);
    \draw [thick] (axis cs:46,-10) -- (axis cs:46,3.0110687733757295);
    \draw [thick] (axis cs:45,-10) -- (axis cs:46,-10,);
    \node at (axis cs:45.6,-11) {\tiny \textit{n.s.}};

    \draw [thick] (axis cs:47,-10) -- (axis cs:47,-0.05747146205524937);
    \draw [thick] (axis cs:48,-10) -- (axis cs:48,-0.0250250421794064);
    \draw [thick] (axis cs:47,-10) -- (axis cs:48,-10,);
    \node at (axis cs:47.6,-11) {\tiny \textit{n.s.}};

    \draw [thick] (axis cs:49,-10) -- (axis cs:49,-7.26937513685628);
    \draw [thick] (axis cs:50,-10) -- (axis cs:50,-2.273698768297811);
    \draw [thick] (axis cs:49,-10) -- (axis cs:50,-10,);
    \node at (axis cs:49.6,-11) {\tiny \textit{n.s.}};

    \draw [thick] (axis cs:51,-10) -- (axis cs:51,-3.895740105961988);
    \draw [thick] (axis cs:52,-10) -- (axis cs:52,0.40769890823004995);
    \draw [thick] (axis cs:51,-10) -- (axis cs:52,-10,);
    \node at (axis cs:51.6,-11) {\tiny \textit{n.s.}};

    \draw [thick] (axis cs:53,-10) -- (axis cs:53,2.060150333928238);
    \draw [thick] (axis cs:54,-10) -- (axis cs:54,4.667581379038131);
    \draw [thick] (axis cs:53,-10) -- (axis cs:54,-10,);
    \node at (axis cs:53.6,-11) {\tiny *};
    \node at (axis cs:53.6,-12) {\tiny -.59};

    \draw [thick] (axis cs:55,-10) -- (axis cs:55,-1.5195633285488215);
    \draw [thick] (axis cs:56,-10) -- (axis cs:56,1.7126225637594266);
    \draw [thick] (axis cs:55,-10) -- (axis cs:56,-10,);
    \node at (axis cs:55.6,-11) {\tiny **};
    \node at (axis cs:55.6,-12) {\tiny -.78};

    \draw [thick] (axis cs:57,-10) -- (axis cs:57,1.7357335896042014);
    \draw [thick] (axis cs:58,-10) -- (axis cs:58,3.2841323110963745);
    \draw [thick] (axis cs:57,-10) -- (axis cs:58,-10,);
    \node at (axis cs:57.6,-11) {\tiny ***};
    \node at (axis cs:57.6,-12) {\tiny -1.02};

    \draw [thick] (axis cs:59,-10) -- (axis cs:59,-1.9920840157056476);
    \draw [thick] (axis cs:60,-10) -- (axis cs:60,5.507066120198044);
    \draw [thick] (axis cs:59,-10) -- (axis cs:60,-10,);
    \node at (axis cs:59.6,-11) {\tiny ***};
    \node at (axis cs:59.6,-12) {\tiny -1.22};

    \draw [thick] (axis cs:61,-10) -- (axis cs:61,-2.4370213868546076);
    \draw [thick] (axis cs:62,-10) -- (axis cs:62,3.9415835998866187);
    \draw [thick] (axis cs:61,-10) -- (axis cs:62,-10,);
    \node at (axis cs:61.6,-11) {\tiny ***};
    \node at (axis cs:61.6,-12) {\tiny -1.32};

    \draw [thick] (axis cs:63,-10) -- (axis cs:63,-4.944919388990212);
    \draw [thick] (axis cs:64,-10) -- (axis cs:64,2.6718665925553027);
    \draw [thick] (axis cs:63,-10) -- (axis cs:64,-10,);
    \node at (axis cs:63.6,-11) {\tiny ***};
    \node at (axis cs:63.6,-12) {\tiny -1.44};

    \draw [thick] (axis cs:65,-10) -- (axis cs:65,1.755185466329182);
    \draw [thick] (axis cs:66,-10) -- (axis cs:66,7.856518448838564);
    \draw [thick] (axis cs:65,-10) -- (axis cs:66,-10,);
    \node at (axis cs:65.6,-11) {\tiny ***};
    \node at (axis cs:65.6,-12) {\tiny -1.53};

    \draw [thick] (axis cs:67,-10) -- (axis cs:67,-1.7702712193199577);
    \draw [thick] (axis cs:68,-10) -- (axis cs:68,6.120392466024429);
    \draw [thick] (axis cs:67,-10) -- (axis cs:68,-10,);
    \node at (axis cs:67.6,-11) {\tiny ***};
    \node at (axis cs:67.6,-12) {\tiny -1.56};

    \node at (axis cs:50,13) {\includegraphics[width=2mm]{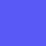} \small SDXL-Turbo White Images};
    \node at (axis cs:50,10.3) {\includegraphics[width=2mm]{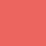} \small SDXL-Turbo Black Images};

  \end{axis}
  \end{tikzpicture}
  \caption{\small Stable Diffusion XL-Turbo exhibits differential White-Black biases, such that generated images of White individuals project as more dominant, electable, and attractive. We plot effect size and significance (* $<$ .05, ** $<$ .01, *** $<$ .001) below each significant comparison.}
  \label{fig:racebias}
\end{figure*}
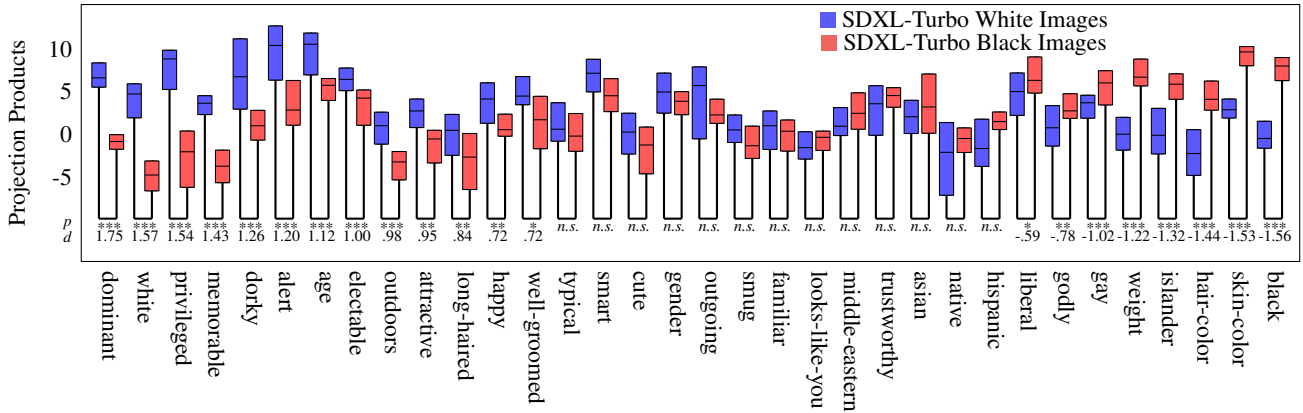

Images generated by Stable Diffusion XL-Turbo also bear signs of racial bias: as shown in Figure \ref{fig:racebias}, we observe statistically significant differences indicating that generated images of White individuals are more likely to be perceived as dominant, privileged, memorable, attractive, electable, and happy than images of Black individuals, which are more likely to be perceived as more liberal and heavy (vs. thin). Note that many of these relationships are \textit{not} observed in the OMI correlation clusters seen in Figure \ref{fig:dendrogram}, indicating that they originate not with the OMI dataset but with the text-to-image model.

\section{Discussion}

Our results make clear the inter-connection of visual perception in AI with the human social world: where a facial impression bias is more consistently shared among humans, CLIP is also more likely to learn it. Facial impression biases have notable consequences in professional and civic life \cite{stoker2016facial}, and prove difficult to dislodge even after intervention \cite{jaeger2020can}. While CLIP may serve as a tool for studying such biases, it may also reinforce or amplify these biases in society, especially given their presence in user-facing image generators.

\subsection{Scale and Bias}
Training on larger datasets results in emergent and amplified facial impression biases. CLIP models exhibit more human-like biases related to trustworthiness and sexuality when trained on LAION-2B, and nearly every OMI attribute increases in model-human similarity between the 80M level and the 400M level. That model parameterization plays no detectable role in facial impression bias underlines that what CLIP models have learned is a biased visual heuristic reflected in the training data, not a more precise representation of an objectively detectable attribute. This also suggests that greater attention to the characteristics of the training data is advisable when pretraining CLIP systems intended for use in downstream applications without fine-tuning, or for supervising other models, including text to image generators using CLIP as a text encoder.

\subsection{Ramifications of Human-like Models}
Amid the excitement over vision-language models like GPT-4 that convincingly imitate aspects of human intelligence \cite{bubeck2023sparks}, the emergence of subtle biases in multimodal models trained on the largest datasets elicits a corrective question: is more ``human-like'' always better? Approximating the distribution of societal associations present in the pretraining dataset as completely as possible may be useful for providing a more general-purpose model. However, as zero-shot vision-language models continue to become more accessible to the general public, including via conversational interfaces that mimic text-based interaction with a human being \cite{li2023blip,liu2024visual}, monitoring for subtle emergent biases in need of mitigation is likely to become a more pressing concern. Humans interacting with fluent, mostly debiased models may in fact be less skeptical of \textit{subtle} reflections of societal bias than of the more blatant and clearly offensive misrepresentations of demographic groups seen in previous generations of models.

\subsection{Implications for Computational Social Science}\label{sec:csss}

The present work is also notable in its consequences for computational social science. By our reckoning, \citet{peterson2022deep} spent tens of thousands of dollars to collect the human subject data needed to learn a supervised model of facial impressions; CLIP produces a model of facial impressions as a side effect of pretraining. While CLIP provides a less precise model of these biases than the supervised counterpart of \citet{peterson2022deep}, our research nonetheless suggests that CLIP models might play a role similar to static word embeddings, which social scientists now routinely employ in computational studies of human attitudes, including in research that generalizes the findings of human subjects experiments \cite{morehouse2023traces,caliskan2022gender}, or which quantifies shifts in human attitudes over decades \cite{borenstein2023measuring,garg2018word}. That CLIP models reflect facial impression bias suggests that they could model other complex sociocultural phenomena not observable via text embeddings alone, a possibility that might be further explored in future work.

\subsection{Limitations}

While participants in the study of \citet{peterson2022deep} were reflective of U.S. demographics as a whole, this also means that a majority of perceivers identified as White, as is clear from the correlation of White and Looks-Like-You OMI attributes in Figure \ref{fig:dendrogram}. Such a demographic skew may render attribute ratings sensitive to correlated biases, given that prior work observes a relationship between social bias and face impressions \cite{xie2021facial}. In addition, while we adopt now-standard prompts specified by OpenAI when introducing CLIP \citet{radford2021learning}, significantly changing these prompts may induce variance in the results.
\section{Conclusion}

The present work demonstrates that three families of CLIP models reflect human facial impression biases, and that human agreement and dataset scale predict how faithfully a CLIP model reproduces a bias. Our findings illustrate the importance of pretraining data for fair and usable zero-shot vision-language models, and they further underscore the interconnection of human bias and AI representations. While concerning for zero-shot applications of CLIP, especially where the model may be used in a user-facing system, our results also demonstrate the remarkable utility of the model for the computational study of societal bias.

\section{Ethical Considerations}

Caution is warranted in a discussion of the findings of this research. While CLIP models trained on larger amounts of data more faithfully encode a human perceptual bias, this does not thereby mean that the models have accurately made an inference about a person based solely on an image of their face. The present work produces no artifact intended for predicting a person's attributes, and \textit{should not be interpreted} as scientific support for the use of vision-language models to predict unobservable attributes. Rather, it should serve as a scientific treatment of emergent social bias arising from large-scale pretraining.

\section{Researcher Positionality}

Our research team includes two individuals with extensive experience in the quantitative study of AI bias and ethics, as well as two individuals with extensive experience in human-computer interaction and value-sensitive design. The team consists of two members who identify as men, and two who identify as women, one of whom is a woman of color. All team members are researchers at the same university in the United States. We sought to include a diversity of perspectives and kinds of expertise given the sensitive subject matter of our work.

\section{Adverse Impacts}

While we have not trained a new model or built any system to facilitate facial analysis, it is possible that an individual may nonetheless attempt to use the methods described in this paper to make unwarranted inferences about individuals based solely on their face. However, these inferences would certainly be less precise than those obtained using a supervised model trained explicitly for facial analysis. Our approach to these potential adverse impacts is to make clear that facial impressions constitute \textit{biases}, and that they intersect with other harmful demographic biases. We hope that demonstrating the presence of these biases in CLIP and Stable Diffusion will serve as a warning of the potential societal effects of using zero-shot vision-language models to classify or represent human beings.

\bibliography{references}

\appendix

\section{Appendix}

\subsection{FaceCLIP Models}

The FaceCLIP models studied in this research are drawn from two sources. The first is the FaRL Github repository at \url{https://github.com/FacePerceiver/FaRL}, which hosts two FaceCLIP models trained from the CLIP-ViT-B-16 architecture for 16 and 64 epochs respectively on the LAION-Face dataset \cite{zheng2022general}. The second is the three Tencent FaceCLIP models available via the Huggingface Hub:

\begin{itemize}
    \item CLIP-ViT-B-32: \url{https://huggingface.co/P01son/FaceCLIP-base-32}
    \item CLIP-ViT-B-16: \url{https://huggingface.co/P01son/FaceCLIP-base-16}
    \item CLIP-ViT-L-14: \url{https://huggingface.co/P01son/FaceCLIP-large-14}
\end{itemize}

\subsection{CLIP Prompts}

The below prompts were used to compute cosine similarities with the images of the OMI dataset.

\subsubsection{Positive Pole Prompts}

\begin{itemize}
\item Trustworthy: a photo of someone who is trustworthy
\item Attractive: a photo of someone who is attractive
\item Dominant: a photo of someone who is dominant
\item Smart: a photo of someone who is smart
\item Age: a photo of someone who is older
\item Gender: a photo of someone who is male
\item Weight: a photo of someone who is overweight
\item Typical: a photo of someone who is typical
\item Happy: a photo of someone who is happy
\item Familiar: a photo of someone who is familiar
\item Outgoing: a photo of someone who is outgoing
\item Memorable: a photo of someone who is memorable
\item Well-groomed: a photo of someone who is well-groomed
\item Long-haired: a photo of someone who has long hair
\item Smug: a photo of someone who is smug
\item Dorky: a photo of someone who is dorky
\item Skin-color: a photo of someone who has dark skin color
\item Hair-color: a photo of someone who has dark hair color
\item Alert: a photo of someone who is alert
\item Cute: a photo of someone who is cute
\item Privileged: a photo of someone who is privileged
\item Liberal: a photo of someone who is liberal
\item Asian: a photo of someone who is asian
\item Middle-eastern: a photo of someone who is middle eastern
\item Hispanic: a photo of someone who is hispanic
\item Islander: a photo of someone who is a pacific islander
\item Native: a photo of someone who is a native american
\item Black: a photo of someone who is black
\item White: a photo of someone who is white
\item Looks-like-you: a photo of someone who looks like me
\item Gay: a photo of someone who is gay
\item Electable: a photo of someone who is electable
\item Godly: a photo of someone who is godly
\item Outdoors: a photo of someone who is outdoors
\end{itemize}

\subsubsection{Negative Pole Prompts}

\begin{itemize}
    \item Trustworthy: a photo of someone who is devious
\item Attractive: a photo of someone who is ugly
\item Dominant: a photo of someone who is subordinate
\item Smart: a photo of someone who is dumb
\item Age: a photo of someone who is young
\item Gender: a photo of someone who is female
\item Weight: a photo of someone who is skinny
\item Typical: a photo of someone who is unusual
\item Happy: a photo of someone who is sad
\item Familiar: a photo of someone who is strange
\item Outgoing: a photo of someone who is shy
\item Memorable: a photo of someone who is forgettable
\item Well-groomed: a photo of someone who is unkempt
\item Long-haired: a photo of someone
\item Smug: a photo of someone who is humble
\item Dorky: a photo of someone
\item Skin-color: a photo of someone who has light skin color
\item Hair-color: a photo of someone who has light hair color
\item Alert: a photo of someone
\item Cute: a photo of someone
\item Privileged: a photo of someone who is disadvantaged
\item Liberal: a photo of someone who is conservative
\item Asian: a photo of someone
\item Middle-eastern: a photo of someone
\item Hispanic: a photo of someone
\item Islander: a photo of someone
\item Native: a photo of someone
\item Black: a photo of someone
\item White: a photo of someone
\item Looks-like-you: a photo of someone who looks like other people
\item Gay: a photo of someone who is straight
\item Electable: a photo of someone
\item Godly: a photo of someone who is sinful
\item Outdoors: a photo of someone who is inside
\end{itemize}

\subsection{Use of AI Assistants}

The first author used Github Co-Pilot when writing programs for this research. AI assistance was used primarily to improve code documentation (\textit{i.e.,} adding docstrings and type-hints), rather than writing the research code itself.

\subsection{Validating Methods}

Following prior work using psychological lexica to validate the performance of language technologies \cite{toney2020valnorm,wolfe2022vast}, we validated our method using the Open Affective Standardized Image Set (OASIS) dataset of \citet{kurdi2017introducing}. OASIS provides ratings of Valence (Pleasantness) and Arousal (Excitement) for 900 images of real-world people, animals, objects, and situations rated by $N$=822 human subjects. Subsequent work also collects human ratings of the Beauty of these 900 images \cite{brielmann2019intense}. We used the formula described in the body of our work to compute the model-human similarity of Valence, Arousal, and Beauty on these images in order to validate the method:

\begin{equation}
    s^{a}_{m} = \rho(\mathbf{m^{a}},\mathbf{h^{a}})
\end{equation}

where the human ratings vector $\mathbf{h^{a}}$ is given by the human Valence, Arousal, or Beauty measurements taken from the datasets above, and the model associations vector $\mathbf{m^{a}}$ is given by:

\begin{equation}
    m^{a}_{j} = cos(\vec{i_{j}}, \vec{t}_{a^{+}}) - cos(\vec{i_{j}}, \vec{t}_{a^{-}})
\end{equation}

Table \ref{tab:vab_ratings} below describes the results of these measurements, which are consistent with validations performed for prior work \cite{wolfe2022hypodescent}.

\begin{table}[]
    \centering
    \begin{tabular}{|l|c|c|c|}
\toprule
Model      &     Valence  &   Arousal &   Beauty \\
\midrule
CLIP RN50           &    0.70*  &   0.21* &   0.63* \\
CLIP RN101       &       0.75*   &  0.25*  &  0.58* \\
CLIP RN50x4        &     0.73*  &   0.23* &   0.40* \\
CLIP RN50x16      &      0.72*   &  0.43*  &  0.59* \\
CLIP RN50x64     &       0.70*   &  0.51*  &  0.42* \\
CLIP ViT-B-32      &     0.67*   &  0.29* &   0.54* \\
CLIP ViT-B-16      &     0.78*   &  0.30* &   0.68* \\
CLIP ViT-L-14      &     0.69*  &   0.53* &   0.50* \\
CLIP ViT-L-14@336px   &  0.71*   &  0.53* &   0.53* \\
\bottomrule
    \end{tabular}
    \caption{\scriptsize OpenAI CLIP models exhibit statistically significant human-model similarity for Valence, Arousal, and Beauty using images of real humans and situations, providing further validation of our methods.}
    \label{tab:vab_ratings}
\end{table}

Note that the positive pole prompts for the valence measurements follow below:

\begin{itemize}
    \item Valence: a photo of something good
    \item Arousal: a photo of something arousing
    \item Beautiful: a photo of something beautiful
\end{itemize}

The negative pole prompts for the above measurements follow:

\begin{itemize}
    \item Valence: a photo of something bad
    \item Arousal: a photo of something dull
    \item Beautiful: a photo of something ugly
\end{itemize}

\subsection{OMI and CLIP Attribute Correlation Matrices}

Below follow the full correlation matrices computed for the OMI dataset and for CLIP-ViT-L-14. The dendrograms for these matrices after hierarchical clustering are provided in the Results section of the paper.

\begin{figure*}
\centering
\includegraphics[width=\textwidth]{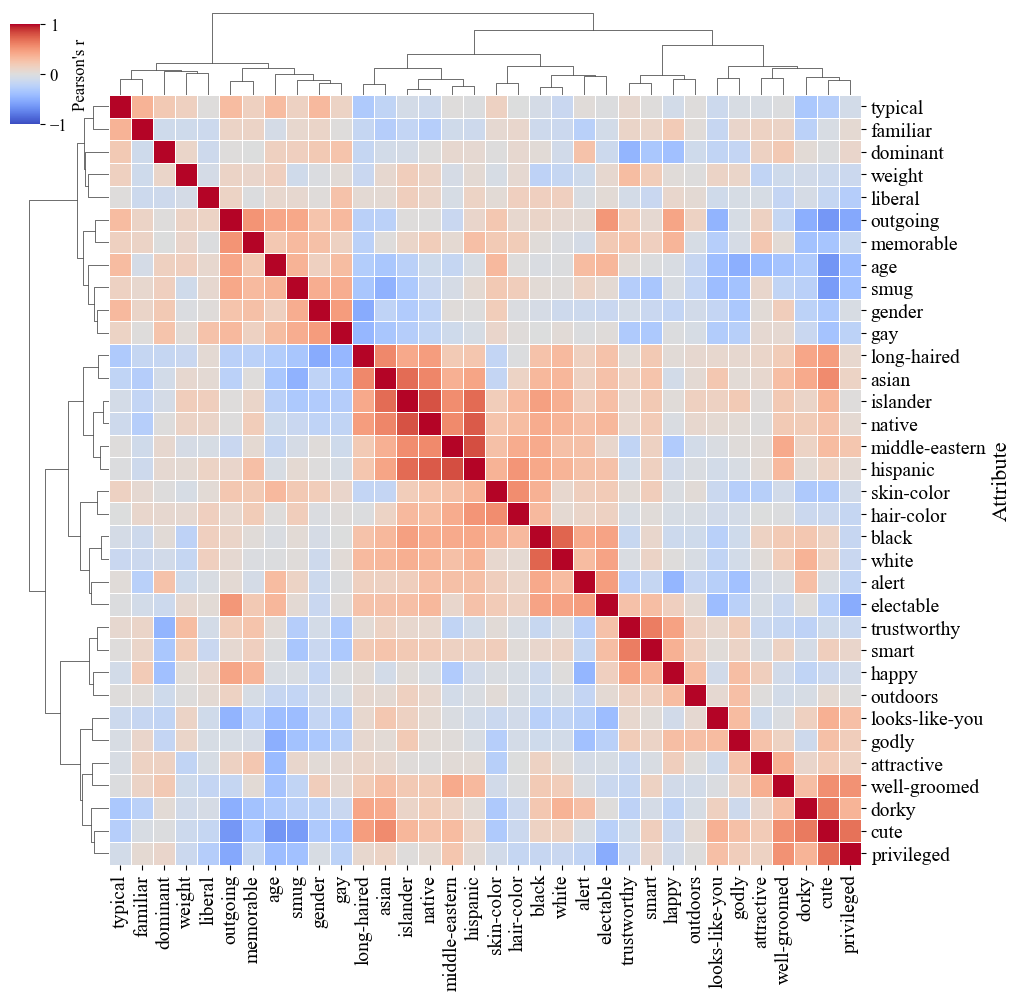}
\caption{\scriptsize The full cross-correlation matrix for OpenAI CLIP-ViT-L-14, the most commonly used CLIP model on the Huggingface Hub \cite{wolf-etal-2020-transformers} and the model with the highest Jaccard similarity to the OMI dataset based on statistically significant correlations.}
\label{cross_correlation_matrix_clip}
\end{figure*}

\begin{figure*}
\centering
\includegraphics[width=\textwidth]{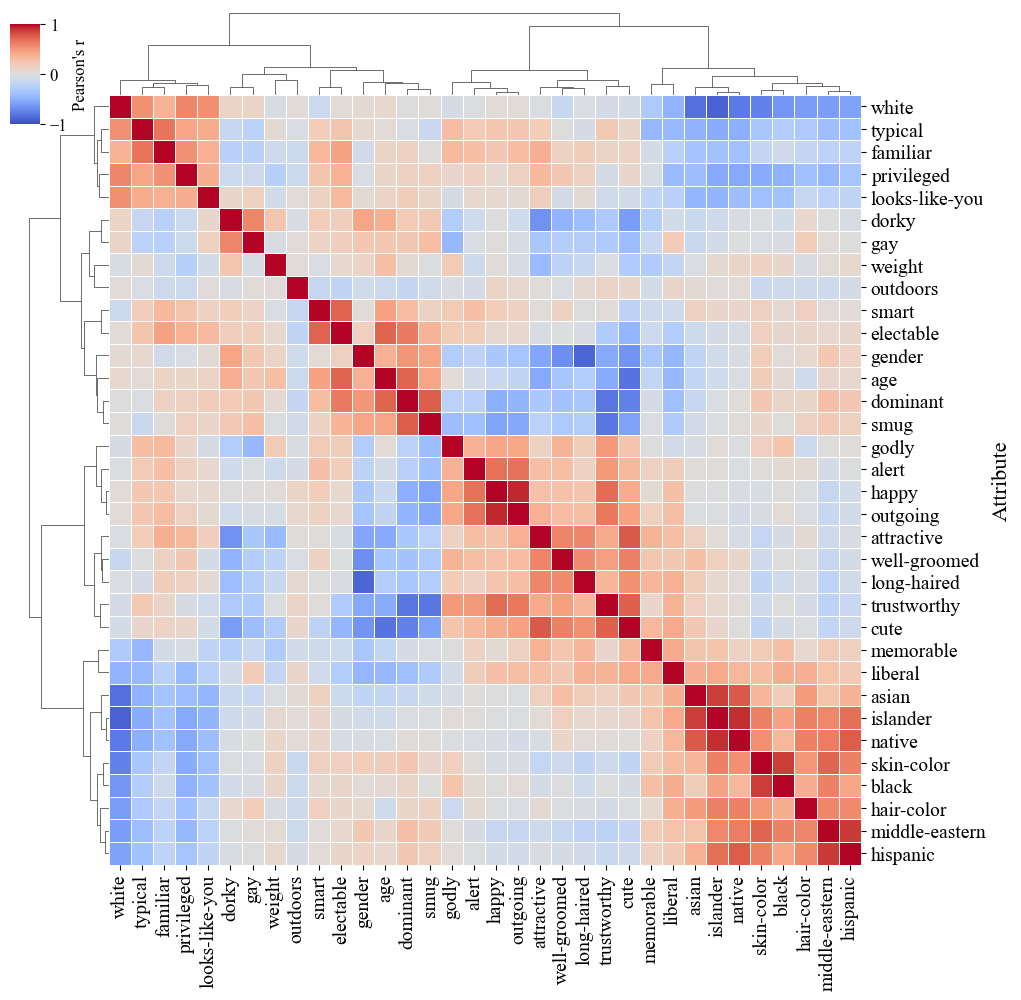}
\caption{\scriptsize The full cross-correlation matrix for OMI dataset, revealing generally stronger and more systematic attribute correlations than in CLIP-ViT-L-14.}
\label{cross_correlation_matrix_omi}
\end{figure*}

\end{document}